\documentclass[10pt,twocolumn,letterpaper]{article}

\usepackage{cvpr}
\usepackage{times}
\usepackage{epsfig}
\usepackage{graphicx}
\usepackage{amsmath}
\usepackage{amssymb}
\usepackage{comment}
\usepackage{bm}
\usepackage{multirow}
\usepackage[dvipdfmx]{}
\usepackage{setspace} 
\newcommand{\stot}{t}%{s\rightarrow t}
\newcommand{\ttos}{s}%{t\rightarrow s}

\usepackage[pagebackref=true,colorlinks,bookmarks=false]{hyperref}
\cvprfinalcopy % *** Uncomment this line for the final submission

\def\cvprPaperID{5190} % *** Enter the CVPR Paper ID here

\ifcvprfinal\fi
\begin{document}

%%%%%%%%% TITLE
%\title{Dual-Cycle Network for Unsupervised Domain Adaptation with Target Shift}
%\title{Partially-Shared Variational Auto-Encoders for \\Unsupervised Domain Adaptation with Imbalance}
\title{Partially-Shared Variational Auto-encoders for \\Unsupervised Domain Adaptation with Target Shift}
\author{
Ryuhei Takahashi\\
Kyoto University\\
{\tt\small takahashi@mm.media.kyoto-u.ac.jp}
\and
Atsushi Hashimoto\\
OMRON SINIC X Corp.\\
{\tt\small atsushi.hashimoto@sinicx.com}
\and
Motoharu Sonogashira\\
Kyoto University\\
{\tt\small sonogashira@mm.media.kyoto-u.ac.jp}
\and
Masaaki Iiyama\\
Kyoto University\\
{\tt\small iiymama@mm.media.kyoto-u.ac.jp}
}

\maketitle
%\thispagestyle{empty}

%%%%%%%%% ABSTRACT
\begin{abstract}
%DeadLine: 11/15
This paper proposes a novel approach for unsupervised domain adaptation (UDA) with target shift. Target shift is a problem of mismatch in label distribution between source and target domains. Typically it appears as class-imbalance in target domain.
In practice, this is an important problem in UDA; as we do not know labels in target domain datasets, we do not know whether or not its distribution is identical to that in the source domain dataset.
Many traditional approaches achieve UDA with distribution matching by minimizing mean maximum discrepancy or adversarial training; however these approaches implicitly assume a coincidence in the distributions and do not work under situations with target shift. 
Some recent UDA approaches focus on class boundary and some of them are robust to target shift, but they are only applicable to classification and not to regression.

To overcome the target shift problem in UDA, the proposed method, partially shared variational autoencoders (PS-VAEs), uses pair-wise feature alignment instead of feature distribution matching. PS-VAEs inter-convert domain of each sample by a CycleGAN-based architecture while preserving its label-related content.
To evaluate the performance of PS-VAEs, we carried out two experiments: UDA with class-unbalanced digits datasets (classification), and UDA from synthesized data to real observation in human-pose-estimation (regression). The proposed method presented its robustness against the class-imbalance in the classification task, and outperformed the other methods in the regression task with a large margin.
\end{abstract}

%%%%%%%%% BODY TEXT
%1．イントロ
\vspace{-1.5em}
\section{Introduction}
% 一段落ずつ，段落の結論を書いていく．
% 他の(優れた)CVPRの文献を参考にすることを推奨(背景や関連研究の書き方が若干日本語の場合と違う．
% 例えばCVPR2017のAppleのやつとか?(3年経つと同じbackgroundでは違和感があるが，構成の参考にはなる)
% https://arxiv.org/abs/1612.07828

\begin{figure}[tb]
\begin{center}
\includegraphics[width=1.0\linewidth]{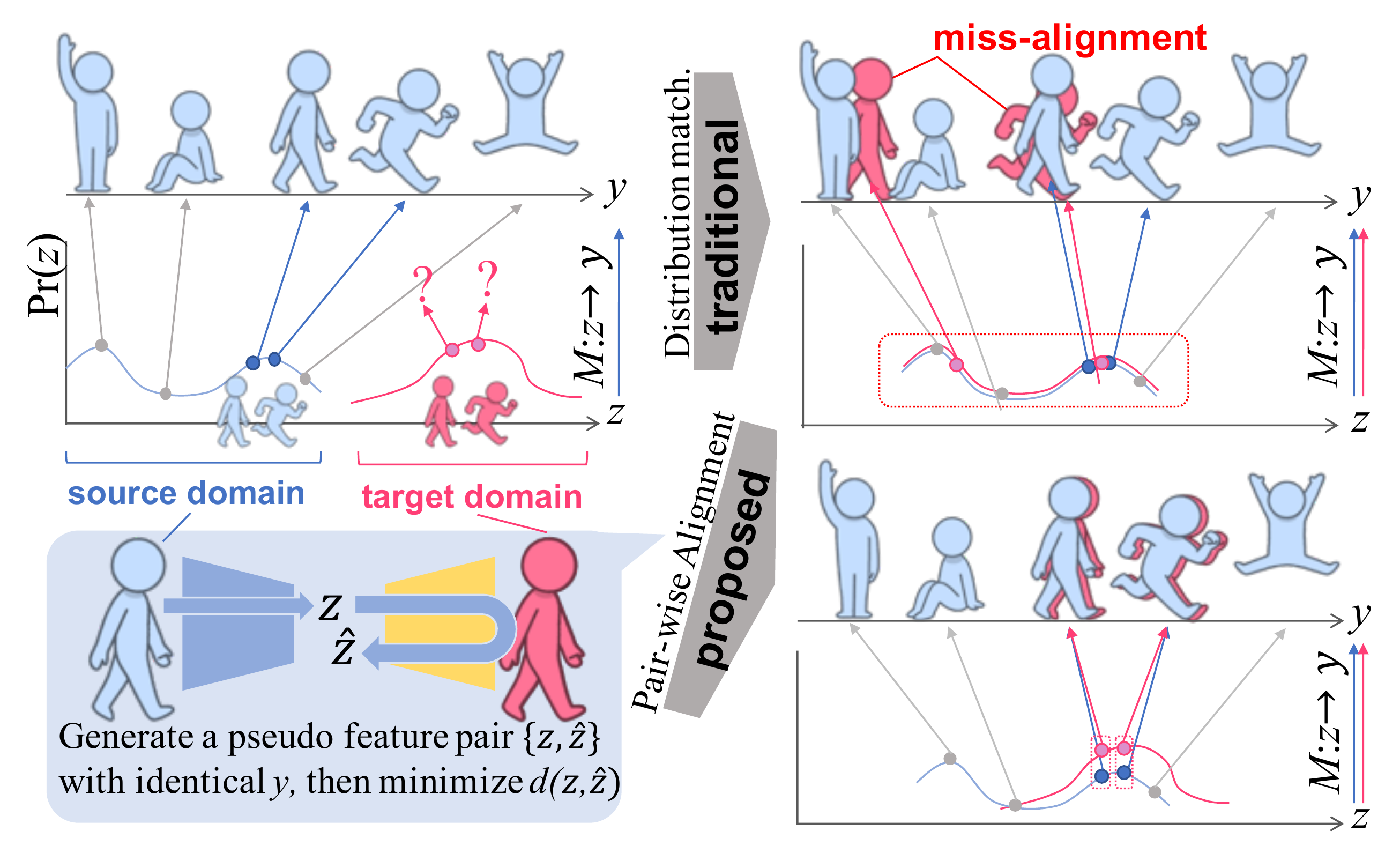}
\caption{(best viewed in color) Overview of the proposed approach. When two domains have different label distributions, feature distribution matching causes miss-alignment in feature space (with classifier/regressor $M:z\rightarrow y$ trained by source domain samples). The proposed method avoids this by sample-wise feature matching, where pseudo sample pairs with identical label are generated via a CycleGAN-based architecture.}
\label{fig:theoretical_overview}
\vspace{-2.0em}
\end{center}
\end{figure}

Unsupervised domain adaptation (UDA) is one of the most studied topics in recent years.
One attractive application of UDA is adaptation from computer graphic (CG) data to sensor-observed (Obs.) data. 
By constructing a CG-rendering system, we can easily obtain a large amount of supervised data with diversity for training.
We denote this task as CG$\rightarrow$Obs. UDA is typically tested on semantic segmentation of traffic scenes \cite{hoffman2017cycada,saito2018maximum} and achieves remarkable performance.

As in ADDA\cite{ADDA}, the typical approach for UDA is to match feature distributions between the source and target domains \cite{dan,dann,ufdn}.
This approach works impressively with balanced datasets, such as those for digits (MNIST, USPS, and SVHN) and traffic-scene semantic-segmentation (GTA5\cite{richter2016playing}$\rightarrow$cityscapes\cite{cordts2016cityscapes}). When the prior label distributions of the source and target domains are mismatched, however, such approaches hardly work without a countermeasure for the mismatch (see Figure \ref{fig:theoretical_overview}).
Cluster finding \cite{saito2018maximum,saito2018adr,cat} is another approach for UDA, and such approaches are more robust against cases of mismatched distributions; they are not, however, applicable to regression problems.
%features from target domain can be easily misaligned by the distribution matching operation when the label distributions of source and target domains are not identical.

In this paper, we propose a novel UDA method applicable to both classification and regression problems with mismatched label distributions.
The problem of mismatched label distributions is also known as target shift \cite{zhang2013domain,gong2016domain} or prior probability shift \cite{moreno2012unifying}. The typical example of this problem is hidden class-imbalance in the target domain dataset.
Previous methods on deep learning models tried to overcome this problem by estimating importance labels for each category  \cite{wdan,Zhang_2018_CVPR,Cao_2018_CVPR,partialDA,Cao_2019_CVPR} or each sample \cite{padaco}. 
The former approach is only applicable to classification tasks, %however <- このhoweverはいらない．英文校正が変．
while the latter under-samples the source domain data.

In contrast, our method resolves this problem by oversampling with data augmentation via the CycleGAN architecture \cite{CycleGAN2017}.
More concretely, the basic strategy of the method is to generate pseudo-pairs of source and target samples (with identical labels) by using the CycleGAN architecture.
Unlike other CycleGAN-based UDA methods \cite{hoffman2017cycada,SBADAGAN}, the proposed method does not match feature distributions. 
Instead, it aligns features extracted from each pseudo-pair in feature space as shown in Figure \ref{fig:theoretical_overview}. 
In addition, the two encoders are designed to share weights. The same encoder is used in encoding any sample of the pseudo-pair. 
Naively minimizing distance between the paired samples leads to bad convergence because of the competition with the losses in CycleGAN training; it implicitly forces  features on the source$\rightarrow$target and target$\rightarrow$source paths to contain different information in addition to the common label-related content.
Hence, we disentangle the features into domain-invariant and domain-specific components to avoid such competition.
We further stabilize training by making the encoders and decoders share weights, giving what we call partially-shared autoencoders (PS-AEs). As a side benefit, this implementation enables us to introduce the mechanism of a variational auto-encoder (VAE) \cite{vae}, which is known to be effective for DA tasks \cite{ufdn,mtan}. % and diverse sample generation \cite{drit}.

%しかし，実際のシーンではこの仮定は成り立つとは限らない．
%$Y_t$のなす分布$\Pr(y \sim Y_t|x \sim X_t)$に，ターゲットデータの取得した環境などの影響によりimbalance（クラス分類ならばあるクラスの画像が多い，姿勢推定ならある特定の姿勢をとった画像が多い，など）が生じるためである．

%% (橋本)蛇足と感じたので↓は省略しました．
%さらに，ターゲットデータのラベルは未知なので，このimbalanceを矯正することは難しい．
%一方で，ソースデータのラベルは存在するものの，やはり$\Pr(y \sim Y_t|x \sim X_t)$は未知なため，$\Pr(y \sim Y_t|x \sim X_t)$を完全に模倣した分布$\Pr(y \sim Y_s|x \sim X_s)$をなすソースデータを用意することもできない．

%このことから，従来のunsupervised domain adaptation は適用可能な場面が限定される．$\Pr(y \sim Y_s) \neq \Pr(y \sim Y_t)$であるような状況下では，$E_s$および$E_t$から得られる$z$を一致させたとしても，共通の推定器$M(y|z)$を用いて得られる分布$\hat{\Pr}(y|\theta_M,\theta_{E_t}) = M(y|z)\Pr(z|x \sim X_t)$は$\Pr(y\sim Y_s)$とは一致するが$\Pr(y \sim Y_t)$とは一致しない．すなわち，特徴分布のマッチングをアイデアとする過去の手法ではデータのimbalanceへの対応が難しい．

%
%ただし，このようにして生成される擬似ペアデータは，$\Pr(y \sim Y_s|x \sim X_s) \neq \Pr(y \sim Y_t|x \sim X_t)$の影響により，必ずしも$y_s = y_t$であるペアにならない．しかし，このモデルではCycleGANにおいて画像再構成を同時に行なっている．そのため，これらの損失関数の重みを調整することにより，それぞれの特徴は完全に一致せず，$\Pr(z|y \sim Y_s)$および$\Pr(z|y \sim Y_t)$であるような$z_s$，$z_t$を抽出できるようになる．

%本研究の主な貢献として、以下が挙げられる．
The contribution of this paper is three-fold.
\begin{itemize}
    \setlength\itemsep{0.1em}
    \setlength\parsep{0.1em}
    %\item ドメイン間でのlabel distributionの不一致という問題に対して，擬似的なペアデータを生成し，そこからドメイン非依存な共通の特徴ベクトルを取り出すという新しいアプローチを提案した．
    \item We propose a novel UDA method that overcomes the target shift problem by oversampling with data augmentation.
    %We firstly proposed a deep-learning-based UDA method that overcomes the target shift problem by oversampling with data augmentation.
    %\item label distributionの不一致な状況下でのUSPS $\rightarrow$ MNIST手書き文字認識において，従来手法よりも高い推定精度を達成した．
    \item The proposed method achieved the best performance for UDA with heavily-class-imbalanced digit datasets.
    %\item 人物姿勢推定で初めて，CGから実写へのUnsupervised Domain Adaptationを実現した．
    \item We tackled the problem of human-pose estimation by UDA with target shift for the first time and outperformed the baselines with a large margin.
\end{itemize}

%2．関連研究
\section{Related Work}

%提案手法の位置付けも書く→ペアワイズな画像変換の話
\begin{table}[tb]
\caption{Representative UDA methods and their supporting situations. The symbol ``(\checkmark)'' indicates that the method theoretically supports the situation but this was not experimentally confirmed in the original paper. 
The abbreviations ``cat.'' and ``reg.'' indicate categorization and regression, respectively.}
\label{tab:comparison}
\begin{tabular}{l|cc|cc}
%\hline
 & \multicolumn{2}{c|}{Balance} & \multicolumn{2}{c}{Imbalance} \\ \cline{2-5} 
 & \!\!cat.\!\!    & \!reg.\!    & \!\!cat.\!\!    & \!reg.\!    \\ \hline
ADDA \cite{ADDA}, UFDN \cite{ufdn}, \!\! &  \multirow{2}{*}{\checkmark}  &  \multirow{2}{*}{(\checkmark)}  &  &               \\
 CyCADA
\cite{hoffman2017cycada}\!\! &    &    &  &               \\
\hline
MCD \cite{saito2018maximum} &  \checkmark  &    & (\checkmark)   &               \\
\hline
PADA \cite{wdan} &  (\checkmark)  &    & \checkmark   &               \\
\hline
SimGAN \cite{shrivastava2017learning} &   &  \checkmark             &                   & (\checkmark)  \\
\hline
\hline
\textbf{Ours} & \checkmark                   &  (\checkmark)            & \checkmark                  & \checkmark  
 %\\ \hline
\end{tabular}
\vspace{-1.5em}
\end{table}

%Table \ref{tab:comparison} shows some representative deep-learning-based methods with its supporting conditions. 
%近年のunsupervised domain  adaptationでは、ターゲット画像から得られる特徴を直接調整することで、ソース画像で学習した識別器で識別可能にする手法が一般的となっている。
The most popular approach in recent UDA methods is to match the feature distributions of the source and target domains so that a classifier trained with the source domain dataset is applicable to target domain samples.
%DAN \cite{dan} is the pioneer work of this approach with deep learning method.
%ターゲット画像から得られる特徴を調整する手法として、入力された特徴がソースとターゲットのどちらからエンコードされたものかを判別するディスクリミネータによる敵対的学習を用いる手法がいくつか発表されている\cite{ADDA,MADDA,dann,partialDA,ufdn}。
There are various options to match the distributions, such as minimizing MMD \cite{dan,wdan}, using a gradient-reversal layer with domain discriminators \cite{dann}, and using alternative adversarial training with domain discriminators \cite{ADDA,MADDA,partialDA,ufdn}.
%このディスクリミネータは、ソース画像の特徴分布とターゲット画像の特徴分布を一致させる役割を持つ。
Adversarial training removes domain bias from the feature representation.
%このようにして特徴分布を一致させた場合、特徴からドメイン固有の情報が失われるため，特徴から元の画像に復元できなくなる．この問題に対処するため，ソースを0、ターゲットを1などとしたようなドメインコードを特徴とともにデコーダに入力し，画像復元を試みる手法もある\cite{ufdn}。
To preserve information in features as much as possible, UFDN \cite{ufdn} adds a decoder to the network for a loss-less encoding. Because the features have no domain information, this method feeds a domain code, a one-hot vector representation for domain reference, to the decoder (with the encoded feature). The encoder and decoder in this model compose a VAE \cite{vae}.
Another approach is feature whitening \cite{FeatureWhitening}, which whitens features from each domain at domain-specific alignment layers. This approach does not use adversarial training, but it tries to analytically fit a feature distribution from each domain to a common spherical distribution. As shown in Table \ref{tab:comparison}, all these methods are theoretically applicable to both classification and regression, but it is limited to the situations without target shift.

MCD was proposed by Saito \etal \cite{saito2018maximum,saito2018adr}, which does not use distribution matching.
Instead, the classifier discrepancy is measured based on the difference of decision boundaries between multiple classifiers. DIRT-T \cite{dirtt} and CLAN \cite{Luo_2019_CVPR} are additional approaches focusing on boundary adjustment.
These approaches are expected to be robust against class imbalance, because they focus only on the boundaries and do not try to match the distributions. 
CAT \cite{cat} is a method that aligns clusters found by other backbone methods. These approaches assume an existence of boundaries between clusters.
Hence, they are not applicable to regression problems, which have continuous sample distributions (see the second row in Table \ref{tab:comparison}).

%本研究の問題設定には、(1)$p(y_s) \neq p(y_t)$である、(2)単純な画像のクラス分類ではなく、相互に依存した複数の出力を持つ回帰問題である、という2つの特徴がある。このうち、(1)に対処する研究として、

%一方，PADA\cite{partialDA}も分布マッチング手法であるが，損失関数に各クラスの出現頻度による重み付けをすることで各ドメイン内の分布$p(y)$の違いを考慮した推定ができる．なお、PADAは$ C_t \subset  C_s  %$(ソースクラス$C_s$にターゲットクラス$C_t$にはないクラスが存在する)という問題設定であり、本研究の問題設定の特別な場合であると言える。
Partial domain adaptation (PDA) is a variant of UDA with several papers on it \cite{wdan,Zhang_2018_CVPR,Cao_2018_CVPR,partialDA,Cao_2019_CVPR} (see the third row in Table \ref{tab:comparison}).
This problem assumes a situation in which some categories in the source domain do not appear in the target domain.
This problem is a special case of UDA with target shift in two senses: it always assumes the absence of a class rather than class-imbalance, and it does not assume a regression task.
The principle approach for this problem is to estimate the importance weight for each category, and ignore those judged as unimportant (under-sampling).
PADACO \cite{padaco} is an extension of PADA for a regression problem and was designed to estimate head poses in UDA with target shift.
It first trains the model with the source domain dataset and obtains pseudo-labels for the target domain dataset. Then, using the similarity of estimated labels, it sets a sampling weight for each sample in the source domain (under-sampling).
Finally, it performs UDA training with a weighted sampling strategy for the source domain dataset.
To obtain better results with this method, it is important to obtain good sampling weights at the first stage. 
In this sense, like CAT \cite{cat}, PADACO requires a good backbone method that provides a good label similarity metric.%, and would boost the accuracy by adjusting the distribution by under-sampling.

Label-preserving domain conversion is another important approach and includes the proposed method (see fourth and fifth rows in Table \ref{tab:comparison}).
Shrivastava \etal proposed SimGAN \cite{shrivastava2017learning}, which converts CG images to nearly real images by adversarial training.
This method tries to preserve labels by minimizing the self-regularization loss, the pixel-value difference between images before and after conversion.
In the sense that the method generates source-domain-like samples from the source domain datasets using GAN, we can say it is a method based on over-sampling with data augmentation. 
We note that this work can be regarded as the first deep-learning-based UDA method for regression that is theoretically applicable to the task with target shift.

%エンコーダを分けてから，パラメータの一部を共有
%・画像変換（Pix2Pix，CycleGAN，StarGAN，etc）
%    ・[教師データがペアになっていればPix2Pixが使える．]
%    ・教師データがペアでないならばCycleGANのような手法が必要となる．
%    ・画像変換を用いて入力実写画像をCGに似せた画像へ変換することで姿勢推定が可能になる．
%    ・画像変換を用いてラベル付きターゲットドメイン画像を生成し，それを教師データとする手法もある（CyCADA）
%
%CycleGANをベースとしてunsupervised domain adaptationを行う手法はいくつか存在している．
%Similar conversion can be achieved via CycleGAN.
%Isola \etal によるCYCADA\cite{hoffman2017cycada}は，ADDA，CycleGAN，Shrivastava \etal の手法の3つを組み合わせて，教師データのソース画像からターゲット画像に似せたものに変換することで、擬似的にラベルを持ったターゲット画像を生成し、それを教師データとして識別器を学習している．

CyCADA\cite{hoffman2017cycada} combines CycleGAN, ADDA and SimGAN for better performance.
It first generates fake target domain images via CycleGAN. The label-consistency of generated samples are preserved by SimGAN's self-regularization loss; however it has a discriminator that matches the feature distributions. 
%It trains target domain classifier with the fake images. At the same time, it matches distribution of features extracted from real samples from source and target domains. %
%しかし，識別器の学習にディスクリミネータによる分布マッチングを行なっているため，従来の分布マッチング手法と同様の問題が生じてしまう．
Hence, this methods principally has the same weakness against target shift.
%また，Russo\etal によるSBADAGAN\cite{SBADAGAN}は，2種類の方法でドメイン適応を行なっている．
SBADAGAN\cite{SBADAGAN} is yet another CycleGAN-based method with discriminator for feature distribution matching.
%classifies target domain samples via ensemble of two UDA classifiers.
%The first classifier is trained with fake target domain images generated by CycleGAN, as well as CyCADA.
%The second classifier is trained with real source domain images, but in the test phase, fake source image converted from target domain samples are fed to it.
%GADAN is designed for a specific purpose of geometry-aware conversion, typically useful for scene text detection/recognition \cite{gadan}.
%一つ目は，CYCADAと同様に擬似的にラベルを持ったターゲット画像を生成してターゲット画像の識別器を学習している．
%二つ目は，入力となるターゲット画像からソース画像に似せたものに変換することで，単にソース画像で学習した識別器での識別を可能にする．
%SBADAGANはこのようにして得られる2つの識別器をアンサンブル的に用いている．

From the viewpoint of human-pose-estimation, a method has been proposed quite recently that estimates human-pose in a UDA manner \cite{zhang2019acmmm}.
It uses a synthesized depth image dataset as the source domain dataset.
The target domain is given with depth and RGB images.
The final goal is to estimate 3D poses from RGB images. 
It performs domain adaptation by transferring knowledge via an additional domain of body-part label representations. The body-part label space are expected to be domain-invariant because of its discrete representation. The method tries to transfer knowledge through this discrete space.
The method was evaluated in UDA, weakly-supervised DA, and fully supervised DA settings.
Target shift was not discussed in this paper because the target domain dataset has enough diversity.
%Because there are a large size target domain dataset for RGB image pose estimation, target shift was not discussed in this paper.

%We note that SimGAN can also deal with categorization problem, but only when two domains are quite similar. 

%We also note that %\cite{cat,padaco} are abbreviated because their performance depends on the backbone method.

%画像変換を用いないドメイン適応手法にも，分布マッチングを用いない手法がある\cite{saito2018maximum,FeatureWhitening}．分布マッチングを行わないこれらの手法により，$p(y_s) \neq p(y_t)$に対処できる可能性がある．

%・ドメイン適応（DANN,UFDN,PADA,PADA2,Taking A Closer Look）
%    ・CG画像に変換するのではなく，実写画像から抽出される特徴を調整することで，直接姿勢推定する手法もある．
%    ・DANNは，ソースドメインで学習した識別器を，ディスクリミネーターを用いてターゲットドメインにも適用可能にする
%    ・ディスクリミネーターは変換元のドメイン（ソースドメイン）と変換先のドメイン（ターゲットドメイン）の分布を一致させる役割を持つ．
%    ・ディスクリミネーターにより，エンコーダがドメイン非依存の特徴のみ抽出するようになると，元の入力画像へ復元できなくなる（画像復元できないとreconstruction-lossが取れない）が，UFDNやStarGANのように，特徴と復元したいドメイン番号をデコーダに同時に入力することで画像復元を試みる手法もある．
%    ・本研究で扱う問題には，1. source domainとtarget domainで$p(y)$が異なる，2. 一つの画像に対して，単一のラベルを推定する識別問題ではなく，複数の間接位置を同時に回帰する問題である．という2つの特徴がある．
%    ・この内，1のような性質に対処する研究として，PADAは各ドメイン内の分布$p(y)$の違いを考慮した推定ができる．
%    ・[PADAはクラス分類問題を対象としている(各クラスにおいて，そのクラスがターゲットドメインに存在するかどうかを推定して重みを付加)．姿勢推定のように，1入力に対して相互に依存した複数の出力が要求されるような問題では重みの調整が難しい．]
%    ・PADAは　Ct（　Cs （ソースクラスCsの内，一部のクラスがターゲットクラスCtに存在しない）という問題設定であり，本研究の問題設定（ターゲットクラスにデータ数の隔たりがある）よりも極端である．
%    ・また，2の特徴に対して，[Taking A Closer LookはセマセグのDA手法であり，ドメイン適応をしたがために逆に推定ミスを起こしてしまうことがある問題を防ぐために，各ピクセルごとにドメイン適応をするかどうかを判断して推定を行う手法である]
%    あと，ちょっと特殊な手法としてMinimum Classifier Discrepancy (CVPR2018)とAdversarial Dropout Regularizationがあるので，これも参照したい．

%%%    
%3．提案手法
\section{Method}
\subsection{Problem statement}
Let $\{x_s,y_s\}\in X_s\times L$ be samples and their labels in the source domain dataset ($L$ is the label space), and let $x_t\in X_t$ be samples in the target domain dataset. The target labels $Y_t$ and their distribution $\Pr(Y_t)$ are unknown (and shiftable from $\Pr(Y_s)$) in the problem of UDA with target shift.
The goal of this problem is to obtain a high-accuracy model for predicting the labels of samples obtained in the target domain.

\subsection{Overview of the proposed method}
\begin{figure*}[!tb]
\begin{center}
\includegraphics[width=1.00\linewidth]{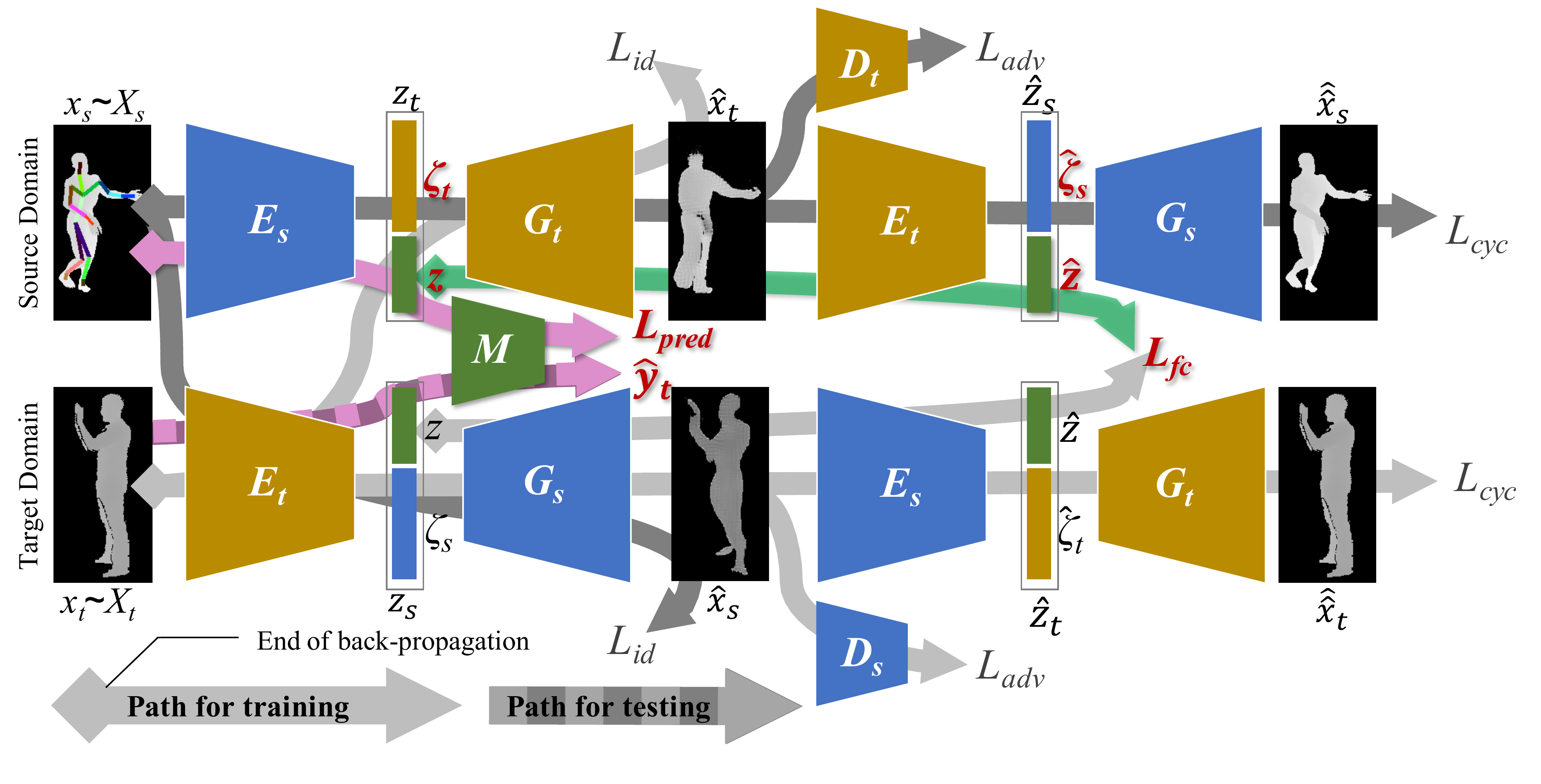} %bb=0 0 1010 514
\vspace{-1.0em}
\caption{(best viewed in color) Architecture of CycleGAN with disentangled features. The major changes from the original CycleGAN (variables, losses with their back-propagating paths) are shown in color.}
\label{fig:cyclegan_arch}
\vspace{-2.0em}
\end{center}
\end{figure*}

The main strategy of the proposed method is to replace the feature distribution matching process with pair-wise feature alignment.
To achieve this, we adopt the CycleGAN architecture shown in Figure \ref{fig:cyclegan_arch}.
The model is designed to generate pseudo pairs $(x_s,\hat{x}_t)$ and $(x_t,\hat{x}_s)$, each of which are expected to have an identical label. 
In addition to CycleGAN's original losses, we add two new losses: $L_{pred}$ for label prediction and $L_{fc}$ for feature alignment, where the losses are calculated only on the domain-invariant component $z$ of the disentangled feature representation $z_s$ (or $z_t$).
After the training, prediction in target domain is done by the path, encoder$\rightarrow$predictor ($M\circ E_t$).
\ref{ss:d-cyclegan} describes this modification in detail.

\begin{figure}[tb]
\begin{center}
\includegraphics[width=1.0\linewidth]{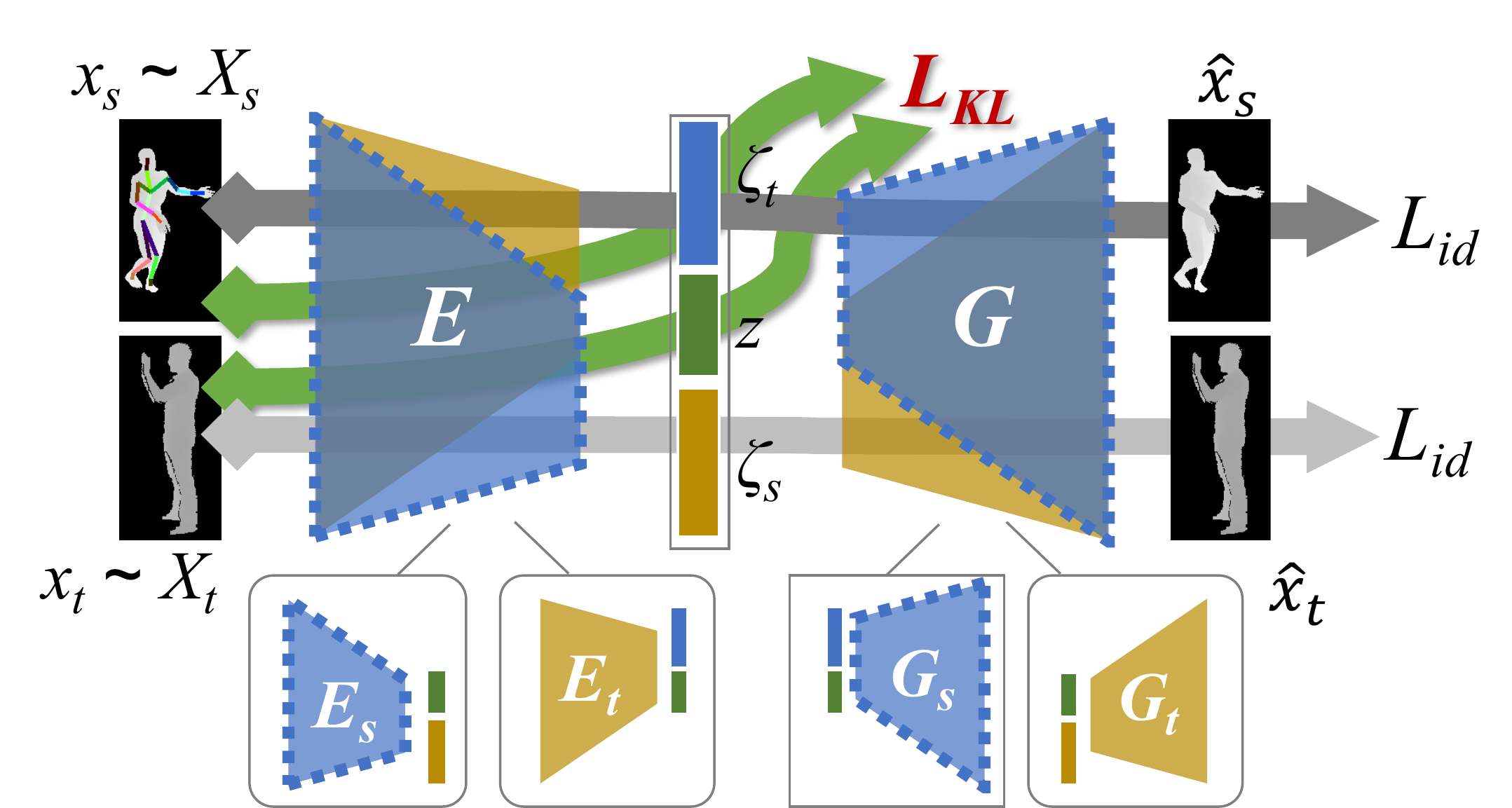} %bb=0 0 621 330
\caption{(best viewed in color) Architecture of partially-shared variational auto-encoders $\hat{x}_s=G_s(E_t(x_s))$ and $\hat{x}_t=G_t(E_s(x_t))$ with the path for calculating $L_{id}$. VAE's re-sampling process is applied when calculating $L_{id}$ but not with other losses.}
\label{fig:psaes}
\vspace{-2.0em}
\end{center}
\end{figure}

To preserve the label-related content at pseudo pair generation, we further modify the network by sharing weights and introducing VAE's mechanism (see Figure \ref{fig:psaes}). \ref{ss:psaes} describes this modification in detail.
%さらに，ペアデータから抽出される特徴が一致するように学習させることで，ドメイン非依存な特徴を取得できるようにする．


\subsection{Disentangled CycleGAN with feature consistency loss}
\label{ss:d-cyclegan}

The model in Figure \ref{fig:cyclegan_arch} has pairs of encoders $E_*$, generators $G_*$, and discriminators $D_*$, where $*\in \{s,t\}$. $\hat{x}_t$ is generated as $\hat{x}_t=G_t(E_s(x_s))$, and $\hat{x}_s$ as $\hat{x}_s = G_s(E_t(x_t))$.
The original CycleGAN \cite{CycleGAN2017} is trained by minimizing the cycle consistency loss $L_{cyc}$, the identity loss $L_{id}$, and the adversarial loss $L_{adv}$ defined in LSGAN \cite{LSGAN}:
\begin{equation}
\vspace{-0.5em}
\!\!\!\!\!
\min_{E_s,E_t,G_s,G_t}\hspace{-1.2em}L_{cyc}(X_s,\!X_t)\!=
\hspace{-0.9em}
\sum_{*\in\{s,t\}}
\hspace{-0.6em}
\mathbb{E}_{x\in X_*}\![d(x,\!G_*(E_{\bar{*}}(\hat{x}_{\bar{*}})))],
\label{eq:l_cyc}
\end{equation}
where $\bar{*}$ is the opposite domain of $*$ and $d$ is a distance function.
\begin{equation}
\vspace{-0.5em}
\min_{E_s,E_t,G_s,G_t}\hspace{-1.2em}L_{id}(X_s,X_t)=\hspace{-0.8em}\sum_{*\in\{s,t\}}\hspace{-0.4em}\mathbb{E}_{x\in X_*}[d(x,G_*(E_*(x)))]
\label{eq:l_rec}
\vspace{-0.5em}
\end{equation}
\begin{equation}
\begin{split}
\hspace{-0.8em}
\min_{E_s,E_t,G_s,G_t}\hspace{-1.0em}&\hspace{1.0em}\max_{D_s,D_t}L_{adv}(X_s,X_t)=\mathbb{E}_{\{x_s,x_t\}\in X_s\times X_t}\\
&[\|D_s(x_s) - 1\|_2+\|D_s(G_s(E_t(x_t))) + 1\|_2\\
&+\|D_t(x_t) - 1\|_2+\|D_t(G_t(E_s(x_s))) + 1\|_2]
\label{eq:l_adv}
\end{split}    
\end{equation}
We note that we used spectral normalization \cite{miyato2018spectral} in $D_s$ and $D_t$ for a stable adversarial training.

To successfully achieve pair-wise feature alignment, the model divides the output of $E_*$ into $z_{\bar{*}}=\{z, \zeta_{\bar{*}}\}$.
Then, it performs feature alignment by using the domain-invariant feature consistency loss $L_{fc}$, defined as
\begin{equation}
\!\!
\min_{E_s,E_t,G_s,G_t}
\hspace{-1.2em}
L_{fc}(Z_s,Z_t)\!\!=
\hspace{-0.8em}
\sum_{*\in\{s,t\}}\!\!\!\!\mathbb{E}_{z\in Z_*}[d(z,\!E_{\bar{*}}(G_{\bar{*}}(z)))],
\label{eq:l_fc}
\end{equation}
where $Z_*=E_*(X_*)$.
Note that gradients are not further back-propagated to $E_*$ over $z$ (see the path of $L_{fc}$ in Figure \ref{fig:cyclegan_arch}) because updating both $z$ and $\hat{z}$ in one step leads to bad convergence.
%This can be seen as the reconstruction loss of $z$'s auto-encoder $E_*\circ G_*$.

In addition, $z$ obtained from $x_s$ is fed into $M$ to train the classifier/regressor $M:z\rightarrow \hat{y}$ by minimizing the prediction loss $L_{pred}(X_s,Y_s)$. 
The concrete implementation of $L_{pred}$ is task-dependent.

We avoid applying $L_{fc}$ to the whole feature components $z_{\stot}$, as it can hardly reach good local minima because of the competition between the pair-wise feature alignment (by $L_{fc}$) and CycleGAN (by $L_{cyc}$ and $L_{id}$). Specifically, training $G_t$ to generate $\hat{x}_t$ must yield a dependency of $\Pr(z_{\stot}|x_t)$. 
This means that $\hat{z}_{\stot}$ is trained to have in-domain variation information for $x_t$.
The situation is the same with $x_s$ and $z_\ttos$.
Hence, $\hat{z}_{\stot}$ and $\hat{z}_{\ttos}$ have dependencies on different factors, $x_t$ and $x_s$, respectively, and it is difficult to match the whole features, $\hat{z}_{\stot}$ and $\hat{z}_{\ttos}$. The disentanglement into $z$ and $\zeta_*$ resolves this situation. Note that this architecture is similar to DRIT \cite{Lee_2018_ECCV}.

\begin{comment}
\begin{equation}
    z_{\stot} = \{z,\zeta_t\},~
    z_{\ttos} = \{z,\zeta_s\}.
\end{equation}
\end{comment}

\begin{figure}[tb]
\begin{center}
\hspace{-2.5em}
\includegraphics[width=1.05\linewidth]{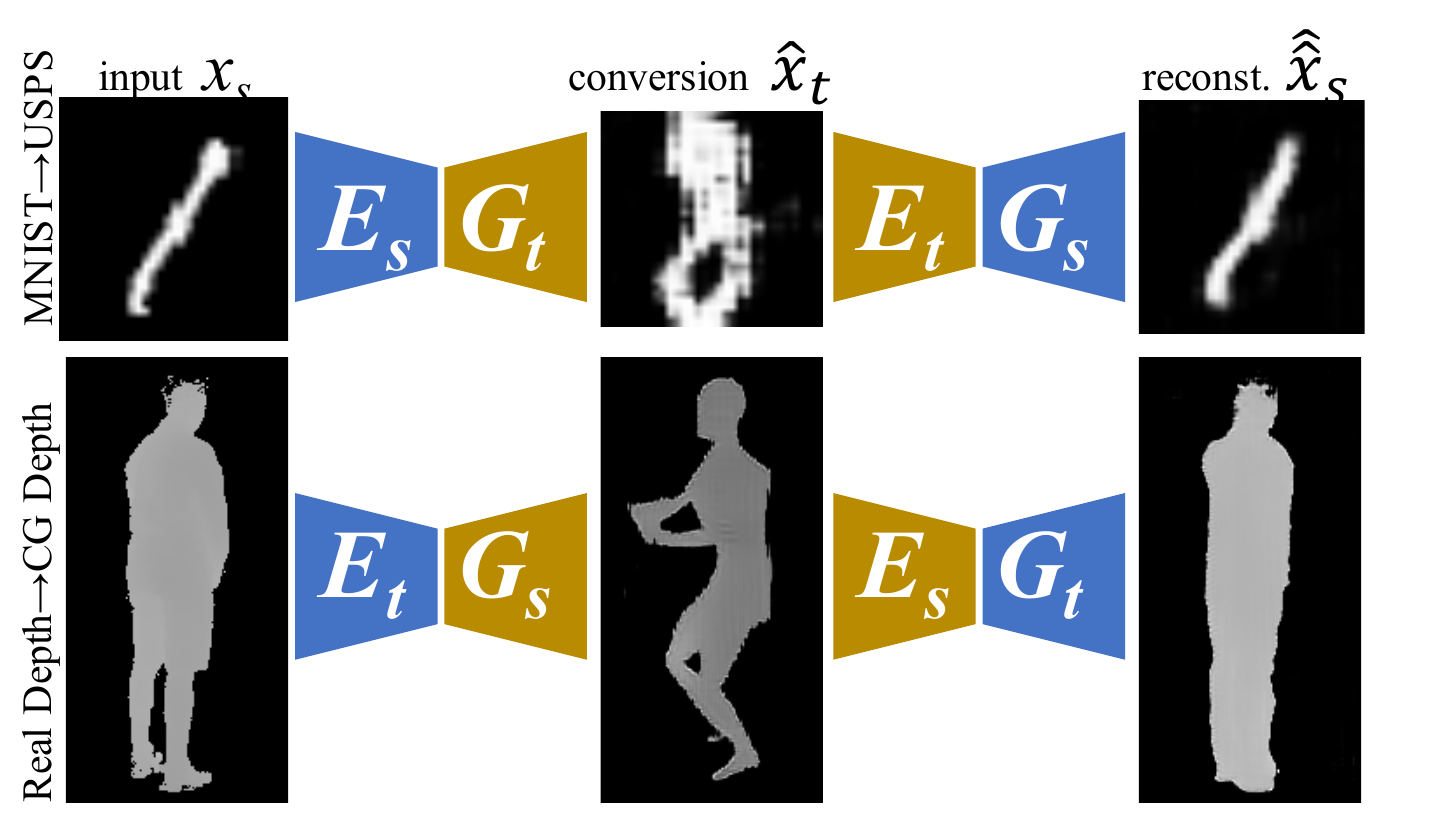}
\caption{Misalignment caused by CycleGAN's two image-space discriminators. This is typically seen with a model that does not share encoder weights.}
\label{fig:misaligned_examples}
\vspace{-2.0em}
\end{center}
\end{figure}

\subsection{Partially shared VAEs}\label{ss:psaes}

Next, we expect $E_s$ and $E_t$ to output a domain-invariant feature $z$. Even with this implementation, however, CycleGAN can misalign an image's label-related content in domain conversions under a severe target shift, because it has discriminators that match not feature- but image-distributions.
Figure \ref{fig:misaligned_examples} shows examples of misalignment caused by image space discriminators.
This happens because the decoders $G_s$ and $G_t$ can convert identical $z$s into different digits, for example, to better minimize $L_{adv}$ with imbalanced distributions. In such cases, the corresponding encoders also extract identical $z$s from images with totally different appearance.
% We note that the identity loss cannot prevent this dexterous conversion because it conditions to generate identical sample by the combination of $\{z,\zeta_*\}$ rather than single $z$.
% Even when $z$ is differently embedded by $E_t(x_s)$, $G_s$ can refer $\zeta_s$ to reconstruct $x_s$ identically.

To prevent such misalignment and get more stable results, we make the decoders share weights to generate similar content from $z$, and we make the encoders extract $z$ only from similar content.
Figure \ref{fig:psaes} shows the details of the parameter-sharing architecture, which consists of units called \textit{partially-shared auto
-encoders} (PS-AEs).
Formally, the partially shared encoders are described as a function $E:x\rightarrow\{z,\zeta_s,\zeta_t\}$. In our implementation, only the last layer is divided into three parts, which outputs $z$, $\zeta_s$, and $\zeta_t$.
$E$ can obviously be substituted for $E_s$ and $E_t$ by discarding $\zeta_t$ and $\zeta_s$ from the output, respectively.
%パラメータを減らすために，E,Gを共有
%$E : x \rightarrow z + \zeta_s + \zeta_t$ 
Similarly, the generator $G:\{z,\zeta_s,\zeta_t\}\rightarrow \hat{x}$ shares weights other than for the first layer, which consists of three parts, which output $z$, $\zeta_s$, and $\zeta_t$. $G$ can be substituted for $G_s$ and $G_t$ by inputting $\{z, \zeta_s, \bm{0}\}$ and $\{z, \bm{0}, \zeta_t \}$, respectively.

%また，ペアデータを生成するための，特徴から画像を生成するジェネレータ$G$もエンコーダと同様にドメイン共通とし，以下のように定式化する．
%$G : z + \zeta_* \rightarrow \hat{x}_*$ 
%ただし，$*$には$s$または$t$のどちらかが入る．このモデルにおいて$\hat{x}_t$を生成する場合，$x_s$を$E$に入力して出力された3つの特徴$z_s$，$\zeta_s$， $\zeta_t$のうち，ドメイン非依存特徴$z_s$とターゲットドメインのドメイン依存特徴$\zeta_t$を$G$に入力することで得られる．
%同様のことが$\hat{x}_s$にも言える．
%すなわち，$\hat{x}_t$および$\hat{x}_s$は，これらを用いて，$\hat{x}_t = G(E(x_s)[z_s,\zeta_t])$， $\hat{x}_s = G(E(x_t)[z_t,\zeta_s])$により得られる．ただし，$f(x)[a]$は，$x$を入力とした$f$の出力のうち，$a$に当たるもののみを取り出すことを意味する．また，このモデルでのcycle lossは以下のように定式化される．

%$\min_{E,G}L_{cyc}=\mathbb{E}_{x_s,x_t}[d(x_s,G(E(\hat{x}_t)[z_t,\zeta_s]))+d(x_t,G(E(\hat{x}_s)[\hat{z}_s,\hat{\zeta}_t]))]$

%ただし，d(a,b)は距離関数であり，本研究ではSmoothL1距離を用いている．SmoothL1距離は以下のように定式化される．

%$d(a,b) =\left\{\begin{array}{ll}
%\|a - b\|_2 & \textrm{if |a - b| < 1 }\\
%\|a - b\| - 0.5 & \textrm{otherwise}
%\end{array}\right.$
%VAE
%MSE -> dに変える 3.3節

%
%CycleGAN
%Partially-shared network
%smoothl1に変える dual cycle loss
%siamese->latent reconstruction loss
% cycleGANと区別してpartially AEsと呼ぶ（reconstruction lossの後に入れる）
%cycle lossの有無は自由（それは言及する）
%\begin{figure}[tb]
%\begin{center}
%\includegraphics[width=10mm]{figure/fake_encode.pdf}
%\includegraphics[width=10mm]{figure/autoencode.pdf}
%\includegraphics[width=10mm]{figure/estimation.pdf}
%\end{center}
%\end{figure}

This implementation brings another advantage for UDA tasks: it can disentangle the feature space by consisting of two variational auto-encoders (VAEs), $G_s \circ E_t$ and $G_t\circ E_s$ (Figure \ref{fig:psaes}).
Putting VAE in a model to obtain a domain-invariant feature is reported as an effective option in recent domain adaptation studies \cite{ufdn,mtan}.
%さらにこのモデルでは，VAEを構築できる．$E$に$x_s$を入力した時に得られる$z_s$，$\zeta_s$， $\hat{\zeta}_t$のうち，$z_s$と$\zeta_s$を$G$に入力することで，入力$x_s$と同じドメインの画像$\Tilde{x}_s$が得られる．同様の手順で$x_t$から$\Tilde{x}_t$が得られる．すなわち，$\Tilde{x}_s$および$\Tilde{x}_t$は，$E$と$G$を用いて，$\Tilde{x}_s = G(E(x_s)[z_s,\zeta_s])$， $\Tilde{x}_t = G(E(x_t)[z_t,\zeta_t])$により得られる．VAEでは，$x_s$および$x_t$がそれぞれ$\Tilde{x}_s$および$\Tilde{x}_t$と同じ画像になるようにreconstruction lossを導入し，特徴$z_s$，$\zeta_s$，$z_t$，$\zeta_t$が標準正規分布に従うようにKL divergence lossを導入する．これらのlossにより，画像のセマンティクス情報が保持されるため，画像変換時にセマンティクスの変化が起きにくくなる．
%reconstruction lossおよびKL lossは以下のように定式化される．ただし$p(z)$は$z\sim N(0,I)$に従う．
%
%In addition to the CycleGAN training, we put an additional loss for making the architecture to two VAEs while calculating $L_{id}$, which can be seen as reconstruction loss of the partially shared VAEs.
To make PS-AEs a pair of VAEs, we put VAE's resampling process at calculation of $L_{id}$ and add the KL loss defined as
\begin{equation}
\begin{split}
& \min_{E_s,E_t}L_{KL}(X_s,X_t)=\\
& \sum_{*\in\{s,t\}}
 \mathbb{E}_{z,\zeta_*\in E_*(X_*)}[
KL(p_{z}|q_{z})+KL(p_{\zeta_*}|q_{\zeta_*})],
%\min_{E}L_{KL}=\mathbb{E}_{z_s,\zeta_s,z_t,\zeta_t}[KL(q(z_s)\|p(z_s))+KL(q(\zeta_s)\|p(\zeta_s))+KL(q(z_t)\|p(z_t))+KL(q(\zeta_t)\|p(\zeta_t))+)]
\label{eq:l_kl}
\end{split}
\end{equation}
where $KL(p\|q)$ is the KL divergence between two distributions $p$ and $q$, $p_{\zeta_*}$ is the distribution of $\zeta_*$ sampled from $X_*$, and $q_{z}$ and $q_{\zeta_*}$ are standard normal distributions with the same sizes as $z$ and $\zeta_*$, respectively.
%Note that UNIT \cite{unit} is another CycleGAN-based method for image-to-image translation combined with VAEs. The architecture of the proposed model is slightly differs from that of UNIT;
%the proposed model shares all weights other than the parts where input/output $z, \zeta_s$, and $\zeta_t$. In contrast, UNIT shares weights only at the nearest parts where input/output the feature.

%ただし，$KL(p\|q)$は$p$と$q$のKL divergenceを表す．このreconstructioin lossは，CycleGAN\cite{CycleGAN2017}におけるidentity lossと同様の効果を得られる．
%また，siamese lossは$x_s$と$\hat{x_t}$のペアおよび$x_t$と$\hat{x_s}$のペアを用いて以下のように定式化される．siamese lossはドメイン非依存特徴である$z$にのみ適用されることに注意．

%$\min_{E_s,E_t}L_{siamese}=\mathbb{E}(d(E(x_s)[z_s],E(\hat{x}_t)[z_t]))+d(E(x_t)[z_t],E(\hat{x}_s)[z_s]))$

%siamese lossは，エンコーダにペアデータを入力した時に，それぞれから得られる$z$を一致させる役割を持つ．これにより$z$はドメイン非依存な特徴となる．なお，$z$がドメイン非依存特徴となった場合，cycle lossやreconstruction lossによる画像復元を維持するために，$\zeta_s$と$\zeta_t$はそれぞれ自然とソースドメインおよびターゲットドメイン依存特徴となる．

%識別器$M$はこのようにして得られた$z$のみを入力とし，ドメイン依存な特徴である$\zeta$は入力しない．識別器は以下のように定式化される．

%$M : z \rightarrow y$

%識別器$M$はソース画像のみで学習され，損失関数$L_{cls}$は解くタスクによって異なる．
%識別器は以下の$L_pose$により学習される．
%$L_{pose}=\mathbb{E}_{z_s,y_s}(MSE(M(z_s),y_s))$
%なお，提案手法をクラス分類（$y_s$がone-hotベクトル）に適用する場合，以下の損失関数を用いる．
%$L_{cls}=\mathbb{E}_{z_s,y_s}(y_s\log M(z_s))$

Our full model, \textit{partially-shared variational auto-encoders} (PS-VAEs), is trained by optimizing the weighted sum of the all the above loss functions:
%これまでに述べた損失関数を組み合わせると，以下の損失関数となる．
\begin{equation}
L_{total}\!=\!L_{adv}\!+\!\alpha L_{cyc}\!+\!\beta L_{id}\!+\!\gamma L_{KL}\!+\!\delta L_{fc}\!+\!\epsilon L_{pred},
\label{eq:l_total}
\end{equation}
%$\alpha$と$\beta$と$\gamma$と$\delta$はハイパーパラメータである．
where $\alpha$, $\beta$, $\gamma$, $\delta$, and $\epsilon$ are hyper-parameters that should be tuned for each task.
For the distance function $d$, we use the smooth L1 distance \cite{smoothL1}, which is defined as
\begin{eqnarray}
d(a,b) =\left\{\begin{array}{ll}
\|a - b\|_2 & \textrm{if}~|a - b| < 1 \\
|a - b| - 0.5 & \textrm{otherwise}
\end{array}\right.
\end{eqnarray}

%\subsection{Sampling trick for more diverse conversion}
%PS-VAEs*の説明．
\begin{comment}
In the proposed model, the distribution of $\zeta_*$ represents in-domain variation.
Because the space of $\zeta_*$ is shared among those obtained from $X_s$ and $X_t$, when $\Pr(\zeta_* \sim X_s)$ and $\Pr(\zeta_* \sim X_t)$ has non-overlapping parts, the conversion $G_*\circ E_{\bar{*}}:X_{\bar{*}}\rightarrow \hat{X}_*$ cannot yields enough diverse samples of $\hat{X}_*$ that cover $X_*$. $\hat{x}_*$ regardless of input sample domain ($x_s$ or $x_t$). 
To cover such diversity, we can use VAE's property; we can sample pseudo $\zeta_*$ from normal distribution $\mathcal{N}(0,1)$.
Hence, we have an option to replace $\zeta_*$ to $\zeta'_*\sim \mathcal{N}(0,1)$ at domain conversion. 
In this case, we can not force the model to minimize $L_{cyc}$ because some information must be lost by this replacing operation. Hence, with this trick, we set $\alpha=0$.
Even with this setting, our model can convert domains owing to partially shared weights. We refer to the model with this trick by PS-VAEs*.
\end{comment}

%4．評価
\section{Evaluation}
%・ドメイン内分布が違う場合に提案手法の優位性を示す．
%実験1：数字画像クラスタリング
%・mnistをソースドメイン，svhnをターゲットドメインとした実験？
%    ・ターゲットドメインの数字の割合を隔たらせることで，本研究の問題設定を実現する．
%実験2：姿勢推定
%・使用データセット：Poserによる生成（CG），CMU panoptic dataset（実写）

%・各テーブルの解説
%・（ハイパラサーチ）
%・定性的評価による考察（画像をのせる）

%この実験では，多くのドメイン適応論文で試されている、手書き数字画像USPS\cite{usps}、MNIST\cite{lecun-mnisthandwrittendigit-2010}，実写数字画像SVHN\cite{svhn}を用いたドメイン適応問題を行う。
%学習時にはターゲットドメインの正解$y_t$は使用せず、モデルの評価時にのみ使用する。
%ここで、本研究の問題設定に合わせるため、ターゲットドメインの各数字の分布を隔たらせた状態にしておく。具体的には，ターゲットドメインでの数字1の割合を10\%，20\%，30\%，40\%，50\%にした状況でUDAを行う．。10\%の場合はもっともbalancedが取れた問題設定であり，50\%は最もbalanceが崩れた問題設定となる。一般的な実験で用いられるtrainデータセットには、各数字の画像数にわずかな隔たりがある。そこで、imbalanceを厳密に統制するため，この実験では1以外の数字の画像数が同じになるように調整した。imbalanceなデータセットの作成には，元のデータセットから指定した割合になるように画像を無作為に取り出す．例えば，MNISTで数字1の割合が50\%であるデータセットを作る場合，数字1の画像を6300枚，他の数字の画像を700枚ずつ取り出す．各バランスで用いたデータ数を\ref{amount}に示す．従来手法に対しても同じデータセットを用いて実験を行った。
\subsection{Evaluation on class-imbalanced digit dataset}
We first evaluated the performance of the proposed method on standard UDA tasks with digit datasets (MNIST\cite{lecun-mnisthandwrittendigit-2010}$\leftrightarrow$USPS\cite{usps}, and SVHN\cite{svhn}$\rightarrow$MNIST). To evaluate the performance under a controlled situation with class-imbalance in the target domain, we adjusted the rate of samples of class `1' from 10\% to 50\%. When the rate was 10\%, the number of samples was exactly the same among the categories. When it was 50\%, half the data belonged to category `1,' which was the most imbalanced setting in this experiment. Note that the original data had slight differences in the numbers of samples between categories. 
We adjusted these differences by randomly eliminating samples. Table \ref{amount} lists the numbers of samples in each dataset and class-imbalance. 
Because SVHN was used only as a source domain, it had no imbalanced situation. 
Note that USPS had only small numbers of samples (500 to 1000 for each category).
Hence, we over-sampled data from category `1' with data augmentation (horizontal shifts of one or two pixels) to achieve the balance.
For the other datasets, we randomly discarded the samples.

\begin{table}
\centering
\label{amount}
\caption{Number of samples under each condition. SVHN was only used as a source domain and had no imbalanced setting.}
\begin{tabular}{|l|r|r|r|r|r|}
\hline

%SVHN$\rightarrow$MNIST
 & 10\% & 20\% & 30\% & 40\% & 50\% \\ \hline \hline 
USPS (1)   & 500 & 1125 & 1922 & 3000 & 4500 \\ %\hline
USPS (other)  & 500 & 500 & 500 & 500 & 500 \\ \hline
MNIST (1) & 4000 & 4500 & 5400 & 6000 & 6300 \\ %\hline
MNIST (other) & 4000 & 2000 & 1400 & 1000 & 700 \\ \hline
SVHN (1)   & 4000 & - & - & - & - \\ %\hline
SVHN (other)  & 4000 & - & - & - & - \\ \hline

\end{tabular}
\vspace{-1.0em}
\end{table}

In this task, $L_{pred}$ is simply given as the following categorical cross-entropy loss:
%識別器$M$の損失関数$L_{cls}$は以下の式で表される．
\begin{equation}
%$L_{cls}=\min_{E,M}\mathbb{E}_{z_s,y_s}[y_s \log M(z_s)]$
\!\!\min_{E_s,M}L_{pred}(Y_s,X_s)\!=\!\mathbb{E}_{x_s,y_s\in X_s\times L} [ - y_s\log M(E(x_s))]
\end{equation}

\begin{comment}
各digit UDA実験の特徴を以下に示す．
\begin{description}
    \item [\textbf{USPS $\rightarrow$ MNIST}] USPSをソースデータ，MNISTをターゲットデータとしたUDA問題である．USPSとMNISTはドメインの性質が似ているため，比較的簡単なUDA問題と言える．
    \item [\textbf{MNIST $\rightarrow$ USPS}] MNISTをソースデータ，USPSをターゲットデータとしたUDA問題である．USPS $\rightarrow$ MNISTと同様，比較的簡単なUDA問題と言える．なお，USPSはdigitひとつあたりの画像数が500枚から1000枚と少ない．そのため，数字1の画像の割合が50\%であるようなimbalanceなデータセットを作成すると，1以外の数字の画像数が100枚になり極端に少なくなってしまう．そこでUSPSをターゲットドメインにする場合，数字1の画像を左右に1または2ピクセルずらすdata augumentationにより数字1の画像を増やすことでimbalanceデータを生成している．
    \item [\textbf{SVHN $\rightarrow$ MNIST}] SVHNにはRGBの背景があり，データの多様性がMNISTよりはるかに大きい．一部の従来手法では，この多様性の差を小さくするために，MNIST画像の各画素をflipさせるdata augumentationを行っている\cite{hoffman2017cycada,ufdn,SBADAGAN}．提案手法においても同様のdata augumentationをした条件下で実験している．
\end{description}
\end{comment}

%また，本実験において比較した従来手法は以下の通りである
We compared the proposed method with the following baselines:
\begin{description}
    \setlength{\itemsep}{0.1em}
	\setlength{\parskip}{0em}
    \item [\textbf{ADDA}\cite{ADDA} and \textbf{UFDN}\cite{ufdn}] %分布マッチングを行うシンプルな手法であり，imbalanceなデータセットにおいて分布マッチング手法がどれくらいの精度を出せるかを測るのに適した手法である．
    are methods based on simple distribution matching. They are applicable to both classification and regression.
    \item[\textbf{PADA\cite{partialDA}}] is also based on distribution matching, but it estimates an importance weight for each category to deal with class-imbalance.
    %imbalanceなデータセットに対処する手法である．原著ではOffice $\rightarrow$ Office HomeデータセットでのUDAでしか実験されていないため，原著でdigit UDAの精度が示されていない．
    \item[\textbf{SimGAN\cite{shrivastava2017learning}}]
    %生成したfake target画像で直接classifierを学習するため，imbalanceの影響を受けにくいと考えられる．
    is a method based on image-to-image conversion. To prevent misalignment during conversion, it also minimizes changes in the pixel-values before and after conversion by using a self-regularization loss. The code is borrowed from the implementation of CyCADA.
    \item[\textbf{CyCADA}\cite{hoffman2017cycada}] 
    is a CycleGAN-based UDA method. The self-regularization loss is used in this method, too. In addition, it matches feature distributions, like ADDA.
    %CycleGANを用いて擬似ラベルを持ったfake target画像を生成するという点は提案手法と近いが，classifierの学習に分布マッチングを用いているという点で異なる．
    
    \item[\textbf{MCD}\cite{saito2018maximum}] %分布マッチングを行わない手法であり，imbalanceなデータセットにも対処できる可能性がある．
    is a method that minimizes a discrepancy defined by the boundary differences obtained from multiple classifiers. This method is expected to be more robust against class-imbalance than methods based on distribution matching, because it does not focus on the entire distribution shape. On the other hand, this kind of approach is theoretically applicable only to classification but not to regression.
\end{description}

Note that some of recent state-of-the-art methods for the balanced digit UDA task was not listed in the experiment due to their reproducibility problem.\footnote{The authors of \cite{FeatureWhitening} provide no implementation and there are currently no other authorized implementations. Two SBADAGAN\cite{SBADAGAN} implementations were available but it was difficult to customize them for this test and the reported accuracy was not reproducible.}
The detailed implementations (network architecture, hyper-parameters, and so on) of the proposed method and the above methods appears in the supplementary material. 
% [コメント:AH]以下は，supplementary materialに書きません？ちょっとずるいが．両方の結果を載せないのであれば．
% We note that some of the recent methods (\cite{ufdn,hoffman2017cycada}) uses pixel-value flipping augmentation at SVHN$\rightarrow$MNIST task for better adversarial training, and we used this option.
%
%SBADAGANは著者推奨GitHubで実験した結果を掲載する予定．
%表が実験の結果である．なお，SBADAGAN\cite{SBADAGAN}やfeature whitening\cite{FeatureWhitening}には著者実装が公開されておらず，imbalance条件下での精度が不明なため，ここには掲載していない．分布マッチングのみを行うADDAやUFDNでは，どのtaskにおいてもimbalanceなデータセットにでは顕著に精度が低下する．CyCADAはSimGANをベースにしているため，MNIST $\rightarrow$ USPSやUSPS $\rightarrow$ MNISTではimbalanceの影響をあまり受けていない．しかし，ドメインの違いが大きいSVHN $\rightarrow$ MNISTではSimGANのself-regularization lossが働かず，imbalanceへの対処ができなくなっている．MCDは識別境界に着目する手法なので分布マッチングベースの手法よりもimbalanceの影響を受けにくい．一方，擬似ペアデータによる特徴の一致を行う提案手法はimbalanceの影響を受けにくく，精度が低下する度合いは低く抑えられている．その結果，数字1の割合が50\%であるようなimbalanceなデータセットの場合において，提案手法が最も高い精度を達成した．
Tables \ref{tab:mtou},\ref{tab:utom}, and \ref{tab:stom} list the results. The methods based on distribution matching (ADDA and UFDN) were critically affected by class-imbalance in the target domain. 
CyCADA was more robust than ADDA and UFDN for the MNIST$\leftrightarrow$USPS tasks, owing to the self-regulation loss, but it did not work for the SVHN$\rightarrow$MNIST task because of the large pixel-value differences between the MNIST and SVHN samples.
In contrast, the proposed method was more robust against imbalance than the above methods, and it was more accurate than PADA and SimGAN.
As a result, our method achieved the best performance in most imbalance settings, and listed its robustness especially with the heaviest imbalance.

\begin{table}
\centering
\caption{Accuracy in the MNIST$\rightarrow$USPS task. The abbreviation ``Ref." indicates reference scores reported in the original papers.}
\label{tab:mtou}
\begin{tabular}{|l|r||r|r|r|r|r|}
\hline
%MNIST$\rightarrow$USPS
& Ref. & 10\% & 20\% & 30\% & 40\% & 50\% \\ \hline \hline
\!\!Source only\!\!  &- & 71.0 & - & - & - & - \\ \hline
ADDA & 89.4 & 89.8 & 86.9 & 79.3 & 81.8 & 78.5 \\ %\hline
UFDN & 97.1 & \textbf{94.0} & 90.4 & 83.2 & 82.3 & 83.8 \\ %\hline
PADA & - & 75.3 & 77.7 & 79.3 & 77.8 & 80.2 \\ %\hline
SimGAN & - & 72.4 & 86.5 & 84.0 & 84.3 & 76.3 \\ %\hline
CyCADA & 95.6 & 91.8 & 91.0 & 80.3 & 86.4 & 87.6 \\ %\hline
MCD & 94.2 & 91.2 & 90.4 & 79.0 & 78.5 & 80.3 \\ \hline
%Ours &&&&&&\\
%~PS-VAEs & - & \textbf{97.1} & \textbf{94.8} & \textbf{93.4} & \textbf{94.6} & \textbf{92.6} \\ 
%~PS-VAEs* &&&&&&\\
%\hline
Ours & - & 93.9 & \textbf{94.8} & \textbf{93.4} & \textbf{94.6} & \textbf{92.6}  \\ \hline

\end{tabular}
\vspace{-1.8em}
\end{table}

\begin{table}
\centering
\caption{Accuracy in the USPS$\rightarrow$MNIST task. The abbreviation ``Ref." indicates reference scores reported in the original papers.}
\label{tab:utom}
\begin{tabular}{|l|r||r|r|r|r|r|}
\hline
%USPS $\rightarrow$MNIST
& Ref. & 10\% & 20\% & 30\% & 40\% & 50\% \\ \hline
\!\!Source only\!\!  & - & 55.6 & - & - & - & - \\ \hline \hline
ADDA & 90.1 & \textbf{96.0} & 89.0 & 81.5 & 78.9 & 80.5 \\ %\hline
UFDN & 93.7& 93.6 & 81.9 & 79.2 & 72.0 & 69.1 \\ %\hline
PADA & - & 47.9 & 39.2 & 36.0 & 29.8 & 25.2 \\ %\hline
SimGAN & - & 68.3 & 50.2 & 49.9 & 63.8 & 49.3 \\ %\hline
CyCADA & 96.5 & 75.3 & 75.3 & 75.2 & 76.7 & 70.7 \\ %\hline
MCD & 94.1 & \textbf{96.0} & 81.5 & 79.1 & 78.1 & 77.4 \\ \hline
Ours & - & 94.8 & \textbf{94.4} & \textbf{90.8} & \textbf{82.6} & \textbf{82.4} \\ \hline
\end{tabular}
\vspace{-0.5em}
\end{table}

\begin{table}
\centering
\caption{Accuracy in the SVHN$\rightarrow$MNIST task. The abbreviation ``Ref." indicates reference scores reported in the original papers.}
\label{tab:stom}
\begin{tabular}{|l|r||r|r|r|r|r|}
\hline

%SVHN$\rightarrow$MNIST
&Ref. & 10\% & 20\% & 30\% & 40\% & 50\% \\ \hline \hline
\!\!Source only\!\!  & - & 46.6 & - & - & - & - \\ \hline
ADDA & 76.0 & 75.5 & 65.0 & 65.2 & 50.8 & 54.3 \\ %\hline
UFDN & 95.0 & 91.1 & 70.9 & 58.7 & 52.6 & 43.6 \\ %\hline
PADA & - & 30.5 & 39.5 & 37.3 & 36.8 & 36.7 \\ %\hline
SimGAN & - & 61.4 & 52.5 & 57.7 & 51.8 & 49.3 \\ %\hline
CyCADA & 90.4 & \textbf{91.4} & 75.4 & 69.7 & 70.7 & 68.3 \\ %\hline
MCD & 96.2 & 90.3 & \textbf{89.7} & \textbf{80.2} & \textbf{72.0} & 65.3 \\ \hline
%Ours & - & 73.7 & 72.9 & 73.8 & 64.4 & \textbf{68.4} \\ 
%\hline
Ours & - & 73.7 & 72.9 & 73.8 & 64.4 & \textbf{68.4} \\ \hline
\end{tabular}
\vspace{-1.0em}
\end{table}

\subsection{Evaluation on human pose dataset}
We also evaluated the proposed method with a regression task on human pose estimation.
For this task, we prepared a synthesized depth image dataset whose poses were sampled with CMU Mocap \cite{cmu_mocap} and rendered with PoserPro2014 \cite{poser}, as the source domain dataset. Each image had 18 joint positions. In the sampling, we avoided pose duplication by confirming that at least one joint had a position more than 50mm away from its position in any other samples. The total number of source domain samples was 15000. These were rendered with a choice of two human models (male and female), whose heights were sampled from a normal distribution with respective means of 1.707 and 1.579m and standard deviations of 56.0mm and 53.3mm).
For the target dataset, we used depth images from the CMU Panoptic Dataset \cite{CMU_panoptic}, which were observed with a Microsoft Kinect.
We automatically eliminated the background in the target domain data by preprocessing.\footnote{The details of this background subtraction appears in the supplementary material.} Finally, we used 15,000 images were used for training and 500 images were used for the test, after manually annotating the joint positions.
%この実験では，CMU panoptic dataset\cite{CMU_panoptic}に付属している、Kinectにより撮影された実写人物深度画像をターゲットドメイン、PoserPro2014\cite{poser}という3DCGモデル生成ソフトにより生成したCGモデル人物深度画像をソースドメインとしたドメイン適応問題を行う。ターゲットドメインであるCMU panoptic datasetには背景がついており，前処理として背景差分を行っている．このようにして得た15000枚をtrain data，500枚をtest dataとした．なお、CMU panoptic datasetの深度画像には関節位置のアノテーションが付属されていないため、テストデータ500枚に対して手動でアノテーションした．ソースドメインでは，CG画像生成を生成するときに人間が取り得ない姿勢をあらかじめ取り除いた．また，全てのCG画像は少なくとも1つの関節位置が離れるような制約をかけており，多様な姿勢をとったデータセットとなっている．CG画像のアノテーションは人物の18の関節の2次元位置である．これを用いて，各関節に対して中心が1となるヒートマップを生成し，これをground truthとして学習を行う．姿勢推定器$M$の損失関数$L_{pose}$は以下の式で表される．

\begin{figure}[!tb]
\begin{minipage}{0.5\hsize}
\begin{center}
Head Position
\vspace{0.1em}
\end{center}
\end{minipage}%
\begin{minipage}{0.5\hsize}
\begin{center}
Left Foot Position
\vspace{0.1em}
\end{center}
\end{minipage}\\
\begin{minipage}{0.25\hsize}
\begin{center}
CG\\
\includegraphics[width=0.95\linewidth]{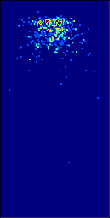}\\ % bb= 0 0 109 218
\end{center}
\end{minipage}%
\begin{minipage}{0.25\hsize}
\begin{center}
Obs.\\
\includegraphics[width=0.95\linewidth]{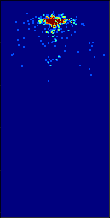}
\end{center}
\end{minipage}%
\begin{minipage}{0.25\hsize}
\begin{center}
CG\\
\includegraphics[width=0.95\linewidth]{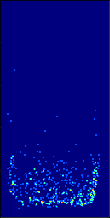} % bb= 0 0 109 218
\end{center}
\end{minipage}%
\begin{minipage}{0.25\hsize}
\begin{center}
Obs.\\
\includegraphics[width=0.95\linewidth]{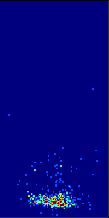}
\end{center}
\end{minipage}%
\caption{Difference in human joint distributions between the source (CG) and target (Obs.) domains. CG images are generated with diverse poses. In contrast, observed images tend to be stagnation.}
\label{fig:pose_differ}
\vspace{-1.0em}
\end{figure}

Figure \ref{fig:pose_differ} shows the target shift between the source and target domains via the differences in joint positions at the head and foot.
In this experiment, we compared the proposed method with SimGAN, CyCADA, and MCD, with an ablation study.
All the methods were implemented with a common network structure, which appears in the supplementary materials.
$L_{pred}$ was defined as
\begin{equation}
    \!\!\min_{E_s,M}L_{pred}(Y_s,X_s)\!=\!\mathbb{E}_{x_s,y_s\in \{X_s,Y_s\}}(d(M(E_s(x_s)),y_s)).
\end{equation}
%\!\!\min_{E_s,M}L_{pred}(Y_s,X_s)\!=\!\mathbb{E}_{x_s,y_s\in X_s\times L} [ - y_s\log M(E(x_s))]
%$L_{pose}=\min_{E_s,M}\mathbb{E}_{z_s,y_s}(d(M(z_s),y_s))$

%また，ソース画像とターゲット画像でとっている姿勢の分布の違いを\ref{fig:pose_differ}に示す．CMU panoptic dataset内の実写人物画像の頭と足の位置の分布は特定の位置に集中している．一方，Poserで生成したデータセット内のCG人物画像の頭と足はより広範囲に分布している．なお，頭と足以外の各関節の位置の分布はsupplementary materialsに掲載している．

\begin{figure}[!tb]
\begin{center}%,bb=0 0 360 218
\includegraphics[width=1.0\linewidth]{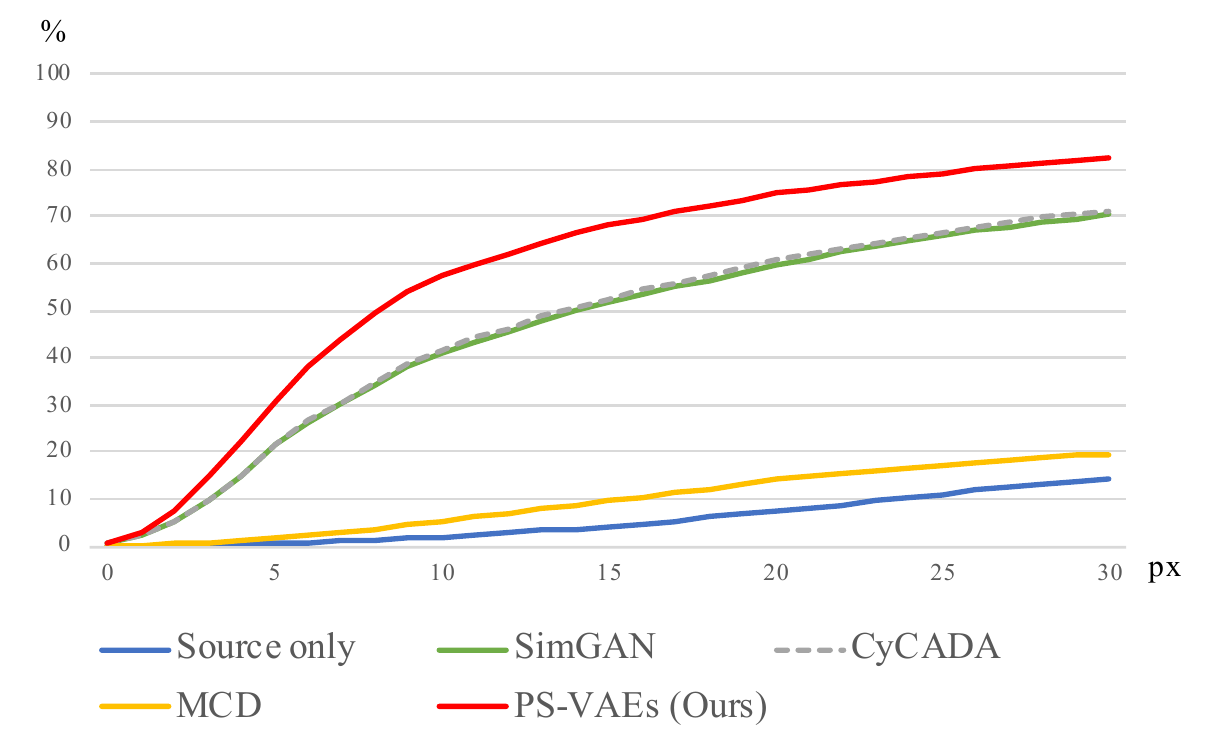}
\caption{(best viewed in color) Averaged percentage of joints detected with errors less than $N$ pixels. (Higher is better.)}
\label{fig:accuracy_per_pixel}
\vspace{-2.5em}
\end{center}
\end{figure}
\begin{table*}
\centering
\caption{Accuracy in human-pose estimation by UDA (higher is better). Results were averaged for joints with left and right entries (e.g., the "Shoulder" column lists the average scores for the left and right shoulders). The "Avg." column lists the average scores over all samples, rather than only the joints appearing in this table.}
%また，ShouldからFootsまでの各関節は左右があるため，表には左右の平均値を載せている．例えば，Shoulderであればright Shoulderとleft Shoulderの平均値を載せている．なお，表で載せている全体の平均値は全ての関節の平均値であり，この表に載せている数値の平均値ではない．
\label{tab:pose_estimation_result}
\begin{tabular}{|l|l|l|l|l|l|l|l|l|l|l|l||l|}
\hline

Error less than 10px. & \!Head\! & \!Neck\! & \!Chest\! & \!Waist\! & \!\!Shoulder\! & \!Elbow\! & \!Wrists\! & \!Hands\! & \!Knees\! & \!Ankles\! & \!Foots\! & \!Avg.\! \\ \hline \hline
Source only & 0.4 & 3.6 & 1.2 & 0.4 & 1.0 & 2.8 & 1.0 & 2.8 & 1.0 & 2.6 & 2.3 & 1.8 \\ %\hline
MCD & 4.6 & 7.0 & 0.2 & 0.6 & 1.4 & 0.2 & 0.3 & 0.9 & 0.4 & 21.0 & 16.6 & 5.3 \\ %\hline
SimGAN& 90.2 & 68.0 & 10.8 & 22.6 & 38.8 & 26.3 & 28.5 & 33.6 & 35.9 & 52.5 & 52.8 & 40.4 \\ %\hline
CyCADA & 90.0 & 69.0 & 15.4 & 28.2 & 39.5 & 27.3 & 31.3 & 32.5 & 35.4 & 54.4 & 53.2 & 41.0 \\ \hline
\hline
Ours &  & 　&  &  &  &  &  &  &  &  &  &  \\ 
~CycleGAN+$L_{fc}$ & 82.8 & 79.0 & 33.8 & 17.0 & 40.0 & 16.4 & 15.8 & 28.4 & 13.8 & 51.0 & 51.5 & 35.5 \\ 
~D-CycleGAN & \textbf{93.0} & \textbf{85.8} & 21.4 & \textbf{47.8} & 42.5 & 42.5 & 35.8 & 39.2 & 42.5 & 66.9 & 64.1 & 50.8 \\  
~D-CycleGAN+VAE &40.6 & 34.2 & 17.6 & 41.2 & 10.1 & 10.2 & 7.5 & 6.4 & 20.0 & 28.0 & 20.2 & 18.6 \\
~PS-AEs & 80.6 & 72.4 & \textbf{40.8} & 28.0 & 46.5 & 28.4 & 25.2 & 29.4 & 25.3 & 58.9 & 53.9 & 42.1 \\ %\hline
~PS-VAEs(full model)\!\! & 89.4 & 84.6 & 21.4 & 43.4 & \textbf{51.7} & \textbf{54.4} & \textbf{49.4} & \textbf{43.9} & \textbf{45.6} & \textbf{74.5} & \textbf{74.0} & \textbf{57.0} \\ 
%~PS-VAEs* &  & 　&  &  &  &  &  &  &  &  &  &  \\ 
\hline
\end{tabular}
\vspace{-0.5em}
\end{table*}

\begin{figure}[!tb]
\vspace{-1.5em}
\begin{minipage}{0.50\hsize}
\begin{center}
\includegraphics[width=\linewidth]{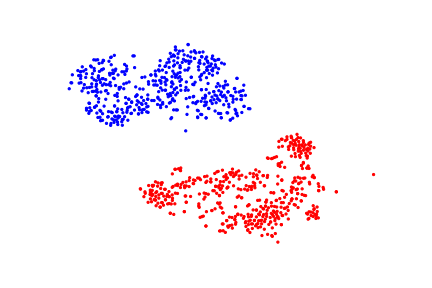}\\
\vspace{-1.0em}
(a) Source Only
\end{center}
\end{minipage}%
\begin{minipage}{0.50\hsize}
\begin{center}
\includegraphics[width=\linewidth]{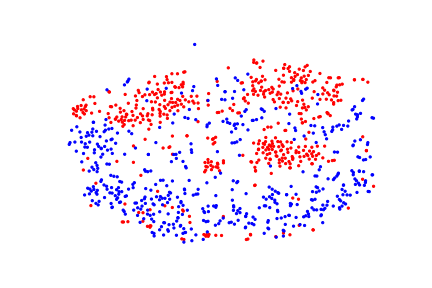}\\
\vspace{-1.0em}
(b) SimGAN
\end{center}
\end{minipage}\\%
\begin{minipage}{0.50\hsize}
\begin{center}
\includegraphics[width=\linewidth]{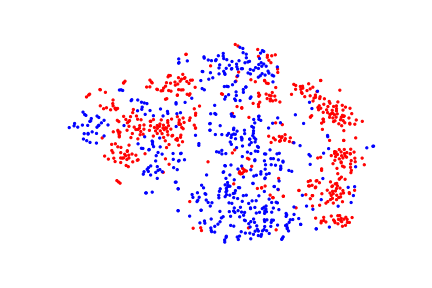}\\
\vspace{-1.0em}
(c) CyCADA
\end{center}
\end{minipage}%
\begin{minipage}{0.50\hsize}
\begin{center}
\includegraphics[width=\linewidth]{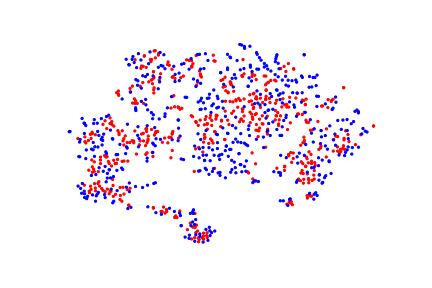}\\
\vspace{-1.0em}
(d) PS-VAEs (Ours)
\end{center}
\end{minipage}
\vspace{0.2em}
\caption{(best viewed in color)  Feature distribution visualized by t-SNE \cite{tsne}: source domain CG data (blue points) and target domain observed data (red points).}
\label{fig:pose_feature_distribution}
\vspace{-1.5em}
\end{figure}
\begin{figure*}[tb]
\begin{minipage}{0.33\hsize}
\begin{center}
\includegraphics[width=1.00\linewidth]{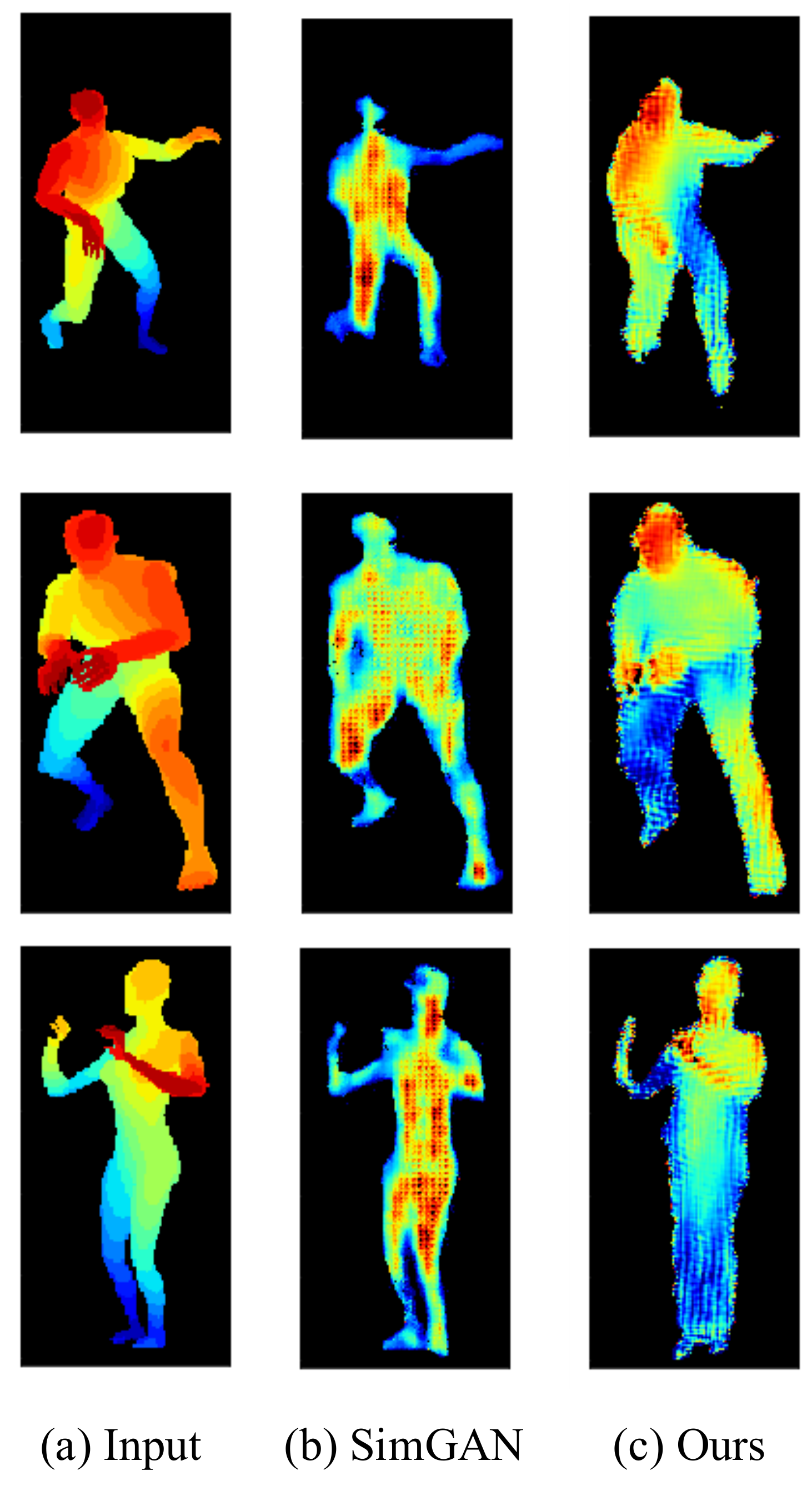}
\vspace{-1.5em}
\caption{(best viewed in color) Qualitative comparison on the domain conversion. Detailed structure in body region is lost with SimGAN, but reproduced with our model.}
\label{fig:pseudo pair results}
\end{center}
\end{minipage}%
\begin{minipage}{0.05\hsize}
~
\end{minipage}%
\begin{minipage}{0.62\hsize}
\begin{center}
\includegraphics[width=1.00\linewidth]{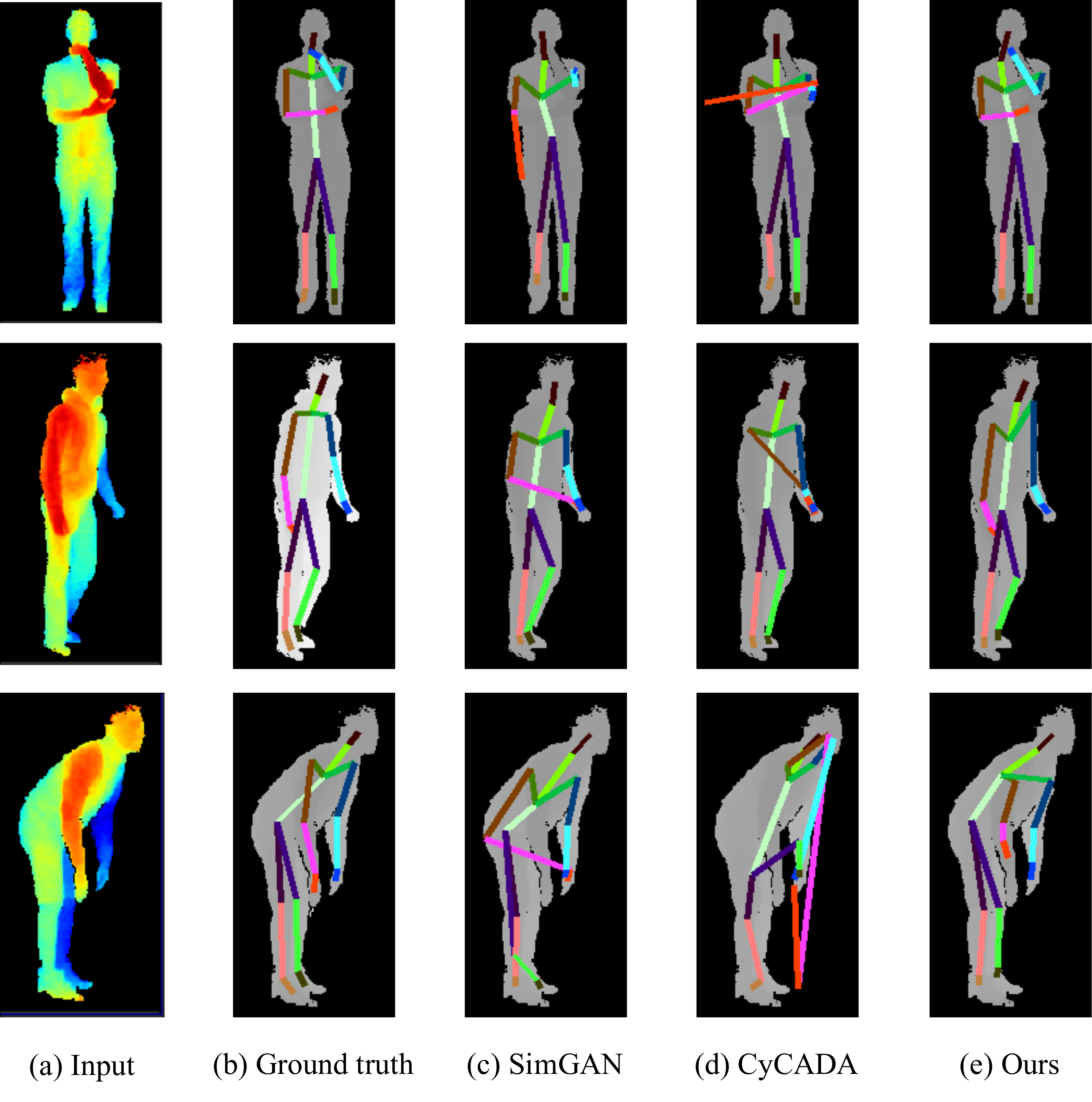}
\caption{(best viewed in color) Qualitative results of human-pose estimation. Due to the lack of detailed depth structure as seen in Fig. \ref{fig:pseudo pair results}, SimGAN and CyCADA often fail to estimate joints with self-occlusion.}
%Note that Input images are normalized for ease of viewing.}
\label{fig:qualitative results}
%\vspace{1.0em} % 
\end{center}
\end{minipage}
\vspace{-1.0em}
\end{figure*}

Figure \ref{fig:accuracy_per_pixel} shows the rate of samples whose estimated joint position error is less than thresholds (the horizontal axis shows the threshold in pixels). Table \ref{tab:pose_estimation_result} lists the joint-wise results in terms of the rate with threshold of ten pixels.
The full model using the proposed methods achieved the best scores on average and for all the joints other than the head, neck, chest, and waist. These four joints have less target shift than others do (see Figure \ref{fig:pose_differ}, for example).
SimGAN was originally designed for a similar task (gaze estimation and hand-pose estimation) and achieved relatively good scores. CyCADA is an extension of SimGAN and has additional losses for distribution matching, but it did not boost the accuracy in the tasks of UDA with target shift. 
MCD was originally designed for classification tasks and did not work for this regression task, as expected.
%表は，各関節において予測の関節位置と正しい関節位置との誤差が10ピクセル以内であるようなテストデータの割合を示している．また，ShouldからFootsまでの各関節は左右があるため，表には左右の平均値を載せている．例えば，Shoulderであればright Shoulderとleft Shoulderの平均値を載せている．なお，表で載せている全体の平均値は全ての関節の平均値であり，この表に載せている数値の平均値ではない．また，\ref{fig:accuracy_per_pixel}は予測と正解の誤差が横軸の数値以内に収まった関節の割合を示している．
%この実験において，我々のPartially-Shared VAEsが最も高い精度を達成した．
%simGANは，目のCG画像を視線推定に利用するために考案されたUDA手法であり，この実験の問題設定に近い．それにより，ある程度高い精度となっている．一方，CyCADAはsimGANに加えて分布マッチングによりclassifierをrefineしている．しかし，ドメイン間でとりうる姿勢の分布が違うことから，refineの効果は微小である．また，MCDはクラス分類やセマンティックセグメンテーションにおいて各クラスの特徴分布の境界を探すようなモデルであり，分布が連続的な回帰問題を解くのが難しいためか精度が低くなっている．
Figure \ref{fig:pose_feature_distribution} shows the feature distributions obtained from four different methods.
%また，擬似ペアデータ生成とdomain consistency lossを用いることでdomain invariantな特徴量を抽出できることをt-SNEにより特徴量を可視化した\ref{fig:pose_feature_distribution}に示す．
%\ref{fig:pose_feature_distribution}(a)ではそれぞれの特徴が完全に分離している．\ref{fig:pose_feature_distribution}(b)では特徴の中にdomain invariantでないものも含まれるため，2つの特徴の混ざり方が不十分である．一方，\ref{fig:pose_feature_distribution}(c)ではそれぞれの特徴が混ざり合い，domain invariantな特徴を抽出できている．
Because SimGAN does not have any mechanisms to align features in the feature space, the distributions did not merge well.
CyCADA better mix the distributions, but still the components are separated.
In contrast, the proposed method merged features quite well despite no discriminators or discrepancy minimization was performed.
This indicates that the proposed pair-wise feature alignment by $L_{fc}$ worked well with this UDA task.

A qualitative difference in domain conversion is shown in Figure \ref{fig:pseudo pair results}.
SimGAN's self-regularization loss worked to keep the silhouette of generated samples, but subtle depth differences in the body regions were not reproduced well.
%The proposed method seems not only to reproduce such subtle depth differences, but also convert domain difference of silhouette; the human model used to synthesize the input does not looks to wear clothes but the generated images seems to wear clothes.
In contrast, the proposed method seems to be able to reproduce such subtle depth differences with the silhouette. This difference contributed to the prediction quality difference shown in Figure \ref{fig:qualitative results}.

%\section{Ablation Study}
%Partially-Shared VAEsの特徴として，CycleGANと違い共通なエンコーダを用いていることと，VAEを導入していることが挙げられる．提案手法では，擬似ペアデータを生成するための画像変換と識別のためのdomain invariantな特徴取得を同時に行う．Partially-Sharedなエンコーダにすることで，Sourceドメインで識別を学習した情報をTargetドメインにも流用できるようになるのでターゲット識別精度の向上が期待できる．また，VAEなしで人物深度画像変換をすると，変換したfake画像は入力画像に近しいシルエットを維持するものの，人物領域内部の微妙な画素値の違いまで表現できていない．しかし，VAEの導入により，このような違いを表現したより高品質なfake画像を生成できる．%画像載せる
In the ablation study, we compared our full model with the following four different variations (see Table \ref{tab:pose_estimation_result}).
\begin{description}
    \item[CycleGAN+$L_{fc}$] does not divide $z_s$ and $z_t$ into the two components, but applied $L_{fc}$ to $z_s$ and $z_t$ directly.
    \item[D-CycleGAN] stands for disentangled CycleGAN, which divides $z_s$ and $z_t$ into the two components, but parameters of encoders and decoders are not shared and not using VAE at the calculation of $L_{id}$. 
    \item[D-CycleGAN+VAE] is a D-CycleGAN with $L_{KL}$ and the resampling trick of VAE at the calculation of $L_{id}$.
    \item[PS-AEs] stands for Partially-shared Auto-Encoders, whose encoders and decoders partially shares parameters as described in \ref{ss:psaes}, but not using VAE.
    \item[PS-VAEs] stands for Partially-shared Variational Auto-Encoders and this is the full model of the proposed method.
\end{description}
D-CycleGAN had actually performed the second best result and D-CycleGAN+VAE and PS-AEs did not work well.
First, as UNIT\cite{unit} does,\footnote{Another neural network model that combines CycleGAN and VAE as the proposed model, but for image-to-image translation.}  it seems to be difficult to use VAE with CycleGAN without sharing weights between the encoder-decoder models.
After combining all these modifications, the full model of the proposed method outperformed any other methods with a large margin.
%While PS-VAEs are more accurate than PS-AEs, D-CycleGAN+VAE is less accurate than D-CycleGAN. Like us, UNIT\cite{unit} that applies VAE also shares weights. There is no proposed method that applies VAE for CycleGAN without sharing weights. On the other hand, it would be difficult to apply VAE without sharing weights at all.

%本研究のメインアイデアは，画像変換を用いて$y_s=y_t$であるような$x_s$と$\hat{x}_t$（または$x_t$と$\hat{x}_s$のペア）を生成して，ペアから抽出されるドメイン非依存な特徴$z_s$と$z_t$を一致させることである．しかし，必ずしも$y_s=y_t$となるペアを生成できるとは限らず，一部のペアでは$y_s \neq y_t$であるようなペアが生成されてしまう．この時，siamese lossがよく機能すると，$y_s \neq y_t$であるにもかかわらず，$z_s=z_t$となってしまう．しかし，このモデルでは$y_s \neq y_t$であるようなペアに対してはあまり機能しない．実際に$y_s=y_t$となったペアと$y_s \neq y_t$となったペアでsiamese lossの値を比較してみると，...
%5．結論
\section{Conclusion}
\vspace{-0.5em}
%・特徴の分離とsiamese-lossにより，ドメイン内分布の違いに影響しない姿勢推定を可能にした．
%本研究では，Partially-Shared Variational Auto Encodersによる擬似ペアデータ生成とfeature consistency lossの導入により，分布マッチングを行うことなくdomain invariantな特徴を得る教師なしドメイン適応手法を提案した．これにより，$p(y_t) \neq p(y_s)$という問題設定の教師なしドメイン適応問題においても精度の低下を抑えられるようになる．また，提案手法はクラス分類問題だけでなく回帰問題に対しても適用可能であり，$p(y_t) \neq p(y_s)$となる代表的な回帰問題である人物姿勢推定の教師なしドメイン適応問題に対しても有効であることを示した．
In this paper, we proposed a novel approach for unsupervised domain adaptation with target shift.
Our approach generates pseudo-feature pairs (with identical labels) to obtain an encoder that aligns target domain samples into the same locations as source domain samples according to their label similarities. Target shift is a common setting in UDA tasks because target domain datasets are often not well organized as the source domain datasets in practice. To be robust against target shift, the method avoids using feature distribution matching but obtains a common feature space by pair-wise feature alignment. To prevent mis-alignment caused at adversarial training in image-space, a CycleGAN-based model was modified to divide features in domain-invariant and domain-specific components, to share weights between two encoder-decoder parts, and to further disentangle features by the mechanism of a variational auto-encoder.
We evaluated the model with digit classification tasks and achieved the best performance under the most of imbalanced situations.
We also applied the method to a regression task of human-pose estimation and found that it outperformed the previous methods significantly.

%-------------------------------------------------------------------------
%\newpage
{\small
\bibliographystyle{ieee}
\bibliography{egbib}
}
\newpage
\appendix

\renewcommand{\thefigure}{\Alph{figure}}
\renewcommand{\thetable}{\Alph{table}}
\twocolumn[
\section{Supplementary Material for Partially-Shared Variational Auto-encoders for Unsupervised Domain Adaptation with Target Shift}
\vspace{2.0em}
]
\subsection{Hyper-parameters}
As defined in Eq. (6) in the main paper,

%Equation \ref{eq:l_t} has five
The proposed method has five
hyper-parameters of $\alpha, \beta, \gamma, \delta, \epsilon$. 
In each experiment, we set these hyper-parameters as listed in Table \ref{tab:hypara}.

%CG画像と実写画像間での各関節の分布の違いを示した図を載せる．

\newcommand{\jointname}{}
\newcommand{\jname}{}
\begin{figure*}[!tb]
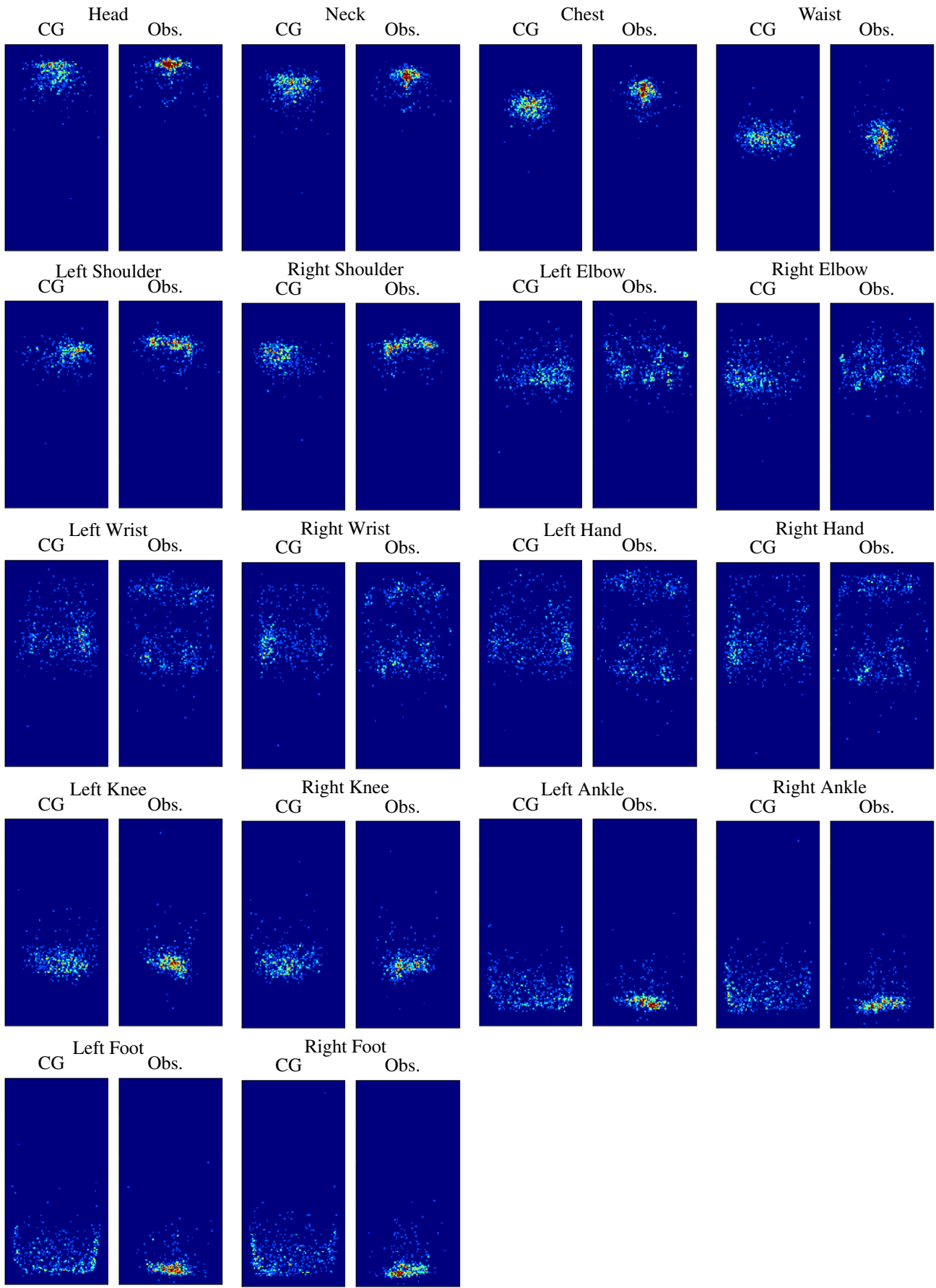

\vspace{-2.0em}
\renewcommand{\jointname}{Head}
\renewcommand{\jname}{Head}
\input{figure/supp_distribution_all_joints_.tex}%
\renewcommand{\jointname}{Neck}%
\renewcommand{\jname}{neck}%
\input{figure/supp_distribution_all_joints_.tex}%
\renewcommand{\jointname}{Chest}%
\renewcommand{\jname}{Chest}%
\input{figure/supp_distribution_all_joints_.tex}%
\renewcommand{\jointname}{Waist}%
\renewcommand{\jname}{Waist}%
\input{figure/supp_distribution_all_joints_.tex}\\
%%%%%%%%
% Left Shoulder Right Shoulder Left Elbow Right Elbow 
\renewcommand{\jointname}{Left Shoulder}
\renewcommand{\jname}{lShoulder}
\input{figure/supp_distribution_all_joints_.tex}%
\renewcommand{\jointname}{Right Shoulder}%
\renewcommand{\jname}{rShoulder}%
\input{figure/supp_distribution_all_joints_.tex}%
\renewcommand{\jointname}{Left Elbow}%
\renewcommand{\jname}{lElbow}%
\input{figure/supp_distribution_all_joints_.tex}%
\renewcommand{\jointname}{Right Elbow}%
\renewcommand{\jname}{rElbow}%
\input{figure/supp_distribution_all_joints_.tex}\\
%%%%%%%%
% Left Wrist Right Wrist Left Hand Right Hand 
\renewcommand{\jointname}{Left Wrist}
\renewcommand{\jname}{lWrist}
\input{figure/supp_distribution_all_joints_.tex}%
\renewcommand{\jointname}{Right Wrist}%
\renewcommand{\jname}{rWrist}%
\input{figure/supp_distribution_all_joints_.tex}%
\renewcommand{\jointname}{Left Hand}%
\renewcommand{\jname}{lHand}%
\input{figure/supp_distribution_all_joints_.tex}%
\renewcommand{\jointname}{Right Hand}%
\renewcommand{\jname}{rHand}%
\input{figure/supp_distribution_all_joints_.tex}\\
%%%%%%%%
% Left Knee Right Knee Left Ankle Right Ankle
\renewcommand{\jointname}{Left Knee}
\renewcommand{\jname}{lKnee}
\input{figure/supp_distribution_all_joints_.tex}%
\renewcommand{\jointname}{Right Knee}%
\renewcommand{\jname}{rKnee}%
\input{figure/supp_distribution_all_joints_.tex}%
\renewcommand{\jointname}{Left Ankle}%
\renewcommand{\jname}{lAnkle}%
\input{figure/supp_distribution_all_joints_.tex}%
\renewcommand{\jointname}{Right Ankle}%
\renewcommand{\jname}{rAnkle}%
\input{figure/supp_distribution_all_joints_.tex}\\
%%%%%%%%%
% Left Foot Right Foot
\renewcommand{\jointname}{Left Foot}
\renewcommand{\jname}{lFoot}
\input{figure/supp_distribution_all_joints_.tex}%
\renewcommand{\jointname}{Right Foot}%
\renewcommand{\jname}{rFoot}%
\input{figure/supp_distribution_all_joints_.tex}%

\caption{Difference in pose distribution between Observed images and CG images.}
\label{fig:posevary}
\vspace{-1.0em}
\end{figure*}

\subsection{Network structure}
%ネットワーク構成：ただし，MNIST$\rightarrow$USPSはResBlockなし．
Table \ref{tab:networks} shows the backbone architecture used in the digit classification task.
We basically used the original backbone architecture for each comparative method if it exists.
In the human pose estimation task, we used the same network architecture designed for this specific task because any other UDA methods have never tried it.

The detail of the backbone network architectures used in the tasks are shown in Table \ref{tab:input}, where ResBlock is defined in Table \ref{tab:resblock}.
In addition to the input data size, they have difference in the architecture of $M$, which is a digit classifier or a human pose regressor.
We also note that ResBlocks in the encoder and the decoder are removed in the task of ResBlock in MNIST$\rightarrow$USPS.
$z$ and $\zeta_*$ are set to 64 and 96 channels in any of the experiments, respectively.
%Our network structure is shown below. In digit classification, we use classifier M. In pose estimation, we use regressor M. Note that there is no ResBlock in MNIST$\rightarrow$USPS.
\subsection{Data augmentation in the SVHN $\rightarrow$ MNIST task}
In the SVHN $\rightarrow$ MNIST task, some conventional methods (UFDN, CyCADA) augmented training images in the MNIST dataset by inverting all pixel values by chance. The reported results of these methods and the proposed method are obtained with this augmentation.
%SVHN \rightarrow MNISTにおいて，一部の従来手法（UFDN， CyCADA）ではMNISTの画素値を反転させた画像を教師データに追加することによるdata augumentationを行っている．提案手法でも同様の画素値を反転したMNIST画像を追加している．%flip画像載せる

\begin{comment}
\begin{figure}[!tb]
\begin{center}%,bb=0 0 360 218
\includegraphics[width=1.0\linewidth]{figure/flip_mnist.pdf}
\caption{MNIST images with flipped pixel values.}
\label{fig:flip}
\vspace{-2.5em}
\end{center}
\end{figure}
\end{comment}

%CMU panoptic datasetは動画になっている．背景は必ず前景よりもdepth値が大きいので，動画の全フレームの中から最大のdepth値を取得すれば背景画像を容易に作成できる．動画を200フレームずつ取り出したものを実写人物画像データとして扱う．背景差分をした後，人物領域を縦横比2:1のbounding boxで切り出し，256 channels×128にリサイズしたものを教師データとして用いた．CGデータには背景がないが，bounding boxによる切り出しとリサイズは同様に行った．

\subsection{Background subtraction for the target domain data in human pose estimation}
The depth images in the CMU panoptic dataset contains cluttered background regions, which are diverse and essentially not related to human pose. These background regions are detected automatically in a pre-process.
Since depth values in a background are always bigger than the foreground in nature, background images can be generated from by obtaining the maximum depth value at each pixel as long as the image data is given as a video captured by a fixed camera.
For all images in the target domain dataset, we obtained its background image in this way and applied background subtraction in advance.
%We got observed human images of 200 frames of those videos.
We set a uniform pixel value to the detected background regions.
After that, the detected foreground regions (i.e., human regions) were cropped with the aspect ratio of 2 : 1 and resized to 256 channels $\times$ 128 pixels. 

Note that the uniform value of the background region is equal to that of CG images. In addition, the same cropping and resizing process were performed to CG images.

\begin{table}
\centering
\caption{hyper-parameters used in the experiments.}
\label{tab:hypara}
\begin{tabular}{|c|rrrrr|}
\hline
 & \multicolumn{1}{c}{$\alpha$} & \multicolumn{1}{c}{$\beta$} & \multicolumn{1}{c}{$\gamma$} & \multicolumn{1}{c}{$\delta$} & \multicolumn{1}{c|}{$\epsilon$} \\ \hline
digit classification & 3 & 1 & 3 & 10 & 3 \\
human pose estimation & 5 & 10 & 3 & 20 & 3 \\ \hline
\end{tabular}
\end{table}
\begin{table}
\centering
\caption{Networks used in each method for digit classification. ``original" indicated that the network is the same with that used in the original paper.}
\label{tab:networks}
\begin{tabular}{|l|l|}
\hline
& digit classification \\ \hline
Source only & same as ADDA   \\ %\hline
ADDA & original \\ %\hline
UFDN & original \\ %\hline
PADA & same as ADDA \\ %\hline
SimGAN & same as CyCADA \\ %\hline
CyCADA & original \\ %\hline
MCD & original  \\ %\hline
PS-VAEs & our digit network \\ \hline
\end{tabular}
\end{table}

\section{Joint distributions}
We show all the distributions of joint positions in Figures \ref{fig:posevary}. Note that right hand can appear in the both sides of the image by chance because the direction of the human is not fixed. 

\begin{table*}
\centering
\caption{Backbone network architecture of digit classification and human pose estimation.}
\label{tab:input}
\begin{minipage}{0.5\textwidth}
\centering
The size of input $x_*$\\
\begin{tabular}{|l|rrr|}
\hline
& height  & width& channel(s) \\ \hline
MNIST$\leftrightarrow$USPS & 32 & 32 & 1  \\ %\hline
SVHN$\rightarrow$MNIST & 32 & 32 & 3 \\ %\hline
pose estimation & 256 & 128 & 1 \\ \hline
\end{tabular}
% 全部　PReLU (Parametrized ReLU)
\\
\vspace{0.5em}
Encoder $E_*$\\
\begin{tabular}{|l|l|}
\hline
%Encoder $E_*$ & \\ \hline \hline
Input& $h\times w\times c$ \\ \hline
Conv& 7 $\times$ 7 $\times$ 64, refrectionpad 3 \\
Parametrized ReLU&\\
Conv& 3 $\times$ 3 $\times$ 128, pad 1, stride 2 \\
Parametrized ReLU&\\
Conv& 3 $\times$ 3 $\times$ 256, pad 1, stride 2 \\
Parametrized ReLU&\\
ResBlock& 256 channels \\
ResBlock& 256 channels \\
ResBlock& 256 channels \\
ResBlock& 256 channels \\
(mu) Conv& 3 $\times$ 3 $\times$ 256, pad 1 \\
(logvar) Conv& 3 $\times$ 3 $\times$ 256, pad 1 \\ \hline
Output & $h/4\times w/4\times 256$ \\
%Output & $h/4\times w/4\times 256$ (for $\mu$ and $\sigma$) \\
\hline
\end{tabular}
\begin{comment}
\begin{table}
\centering
\begin{tabular}{|l|l|}
\hline
Divide Encoder output as follows:\\ \hline
$z : 64$ channels\\
$\zeta_s : 96$ channels\\
$\zeta_t : 96$ channels\\
\hline
\end{tabular}
\end{table}
\end{comment}
\\
\vspace{0.5em}
Decoder $G_*$\\
\begin{tabular}{|l|l|}
\hline
Input& $h/4\times w/4 \times 256$\\ \hline
ResBlock& 256 channels\\
ResBlock& 256 channels\\
ResBlock& 256 channels\\
ResBlock& 256 channels\\
Upsampling& 2 $\times$ 2 $\times$ 256\\
Conv& 3 $\times$ 3 $\times$ 128, pad 1\\
Parametrized ReLU&\\
Upsampling& 2 $\times$ 2 $\times$ 128\\
Conv& 3 $\times$ 3 $\times$ 64, pad 1\\
Parametrized ReLU&\\
Conv& 7 $\times$ 7 $\times c$, reflectionpad 3\\ \hline
Output& $h\times w\times c$ \\ \hline
\end{tabular}
\\
\vspace{0.5em}
Classifier $M$ (for digit classification)\\
\begin{tabular}{|l|l|}
\hline
%Classifier M&\\ \hline \hline
Input& 8$\times$8$\times$64 channels\\ \hline
Flatten&4096\\
Parametrized ReLU&\\
Dropout& 50\% \\
%Full connection& 4096 $\times$ 10 \\ \hline
Full connection& 10 \\ \hline
Output& 10 \\ \hline
\end{tabular}
\end{minipage}%
\begin{minipage}{0.5\textwidth}
\centering
Regresser $M$ (for human pose estimation)\\
\begin{tabular}{|l|l|}
\hline
Regressor M&\\ \hline \hline
Input& 8$\times$8$\times$64 channels\\ \hline
ResBlock& 64 channels\\
ResBlock& 64 channels\\
ResBlock& 64 channels\\
ResBlock& 64 channels\\
Upsampling& 2 $\times$ 2 $\times$ 64\\
Conv& 3 $\times$ 3 $\times$ 32, pad 1\\
Parametrized ReLU&\\
Upsampling& 2 $\times$ 2 $\times$ 32\\
Conv& 3 $\times$ 3 $\times$ 16, pad 1\\
Parametrized ReLU&\\
Conv& 7 $\times$ 7 $\times$ 18, reflectionpad 3\\ \hline
Output& $h\times w\times18$ \\ \hline
\end{tabular}
\\
\vspace{0.5em}
Discriminator $D_*$\\
\begin{tabular}{|l|l|}
\hline
%Discriminator D&\\ \hline \hline
Input& $h\times w\times c$\\ \hline
Conv& 4 $\times$ 4 $\times$ 64, pad 2, stride 2\\
Spectral norm&\\
Parametrized ReLU&\\
Conv& 4 $\times$ 4 $\times$ 128, pad 2, stride 2\\
Spectral norm&\\
Parametrized ReLU&\\
Conv& 4 $\times$ 4 $\times$ 256, pad 2, stride 2\\
Spectral norm&\\
Parametrized ReLU&\\
Conv& 4 $\times$ 4 $\times$ 512, pad 2, stride 2\\
Spectral norm&\\
Parametrized ReLU&\\
Conv& 4 $\times$ 4 $\times$ 512$\times$, pad 2\\
Spectral norm&\\
Parametrized ReLU&\\
Conv& 4 $\times$ 4 $\times$ 1, pad 2\\ \hline
Output& $(h/16 + 4)\times(w/16 + 4)$ \\ \hline
\end{tabular}
\end{minipage}
\end{table*}

\begin{table*}
\centering
\caption{Definition of ResBlock.}
\label{tab:resblock}
\begin{tabular}{|l|l|}
\hline
%ResBlock n channels &\\ \hline \hline
Input& $h\times w \times c_{in}$ \\ \hline
Conv& 3 $\times$ 3 $\times c_{in}$, pad 1 \\
Parametrized ReLU&\\
Conv& 3 $\times$ 3 $\times c_{in}$, pad 1 \\ 
Parametrized ReLU&\\ \hline
Output& $h\times w \times c_{out}$ \\ \hline
\end{tabular}
\end{table*}

\begin{table*}
\centering

\begin{tabular}{|l|l|l|l|l|l|l|l|l|l|l|l||l|}
\hline

Error less than 10px. & \!Head\! & \!Neck\! & \!Chest\! & \!Waist\! & \!\!Shoulder\! & \!Elbow\! & \!Wrists\! & \!Hands\! & \!Knees\! & \!Ankles\! & \!Foots\! & \!Avg.\! \\ \hline \hline
Source only & 0.4 & 3.6 & 1.2 & 0.4 & 1.0 & 2.8 & 1.0 & 2.8 & 1.0 & 2.6 & 2.3 & 1.8 \\ %\hline
MCD & 4.6 & 7.0 & 0.2 & 0.6 & 1.4 & 0.2 & 0.3 & 0.9 & 0.4 & 21.0 & 16.6 & 5.3 \\ %\hline
SimGAN& \textbf{90.2} & 68.0 & 10.8 & 22.6 & 38.8 & 26.3 & 28.5 & 33.6 & 35.9 & 52.5 & 52.8 & 40.4 \\ %\hline
CyCADA & 90.0 & 69.0 & 15.4 & 28.2 & 39.5 & 27.3 & 31.3 & 32.5 & 35.4 & 54.4 & 53.2 & 41.0 \\ \hline
~PS-VAEs(Ours)\!\! & 89.4 & \textbf{84.6} & \textbf{21.4} & \textbf{43.4} & \textbf{51.7} & \textbf{54.4} & \textbf{49.4} & \textbf{43.9} & \textbf{45.6} & \textbf{74.5} & \textbf{74.0} & \textbf{57.0} \\ 
%~PS-VAEs* &  & 　&  &  &  &  &  &  &  &  &  &  \\ 
\hline
\end{tabular}
\vspace{-0.5em}
\end{table*}

\end{document}